\PassOptionsToPackage{dvipsnames,table}{xcolor}
\documentclass[11pt]{article} 
\usepackage[utf8]{inputenc}
\usepackage{geometry}
\usepackage{url,hyperref,lineno,microtype,subcaption}
\usepackage[onehalfspacing]{setspace}

\usepackage[normalem]{ulem}
\usepackage{graphicx}

\usepackage{nicefrac}
\usepackage{longtable}

\usepackage[normalem]{ulem} 

\usepackage{amsmath}
\usepackage{amsfonts}

\DeclareMathOperator*{\argmin}{arg\,min}

\usepackage[dvipsnames]{xcolor}
\newcommand{\mlemph}[1]{{\color{blue}#1}}

\usepackage[symbol]{footmisc}

\newcommand{\revised}[1]{#1}

\begin{document}

\begin{center}
\rule{\textwidth}{2pt}\vspace{5pt}
    {\noindent \Large The Impact of Machine Learning on 2D/3D Registration for Image-guided Interventions: A Systematic Review and Perspective}
    
    \vspace{10pt}
    
    {\noindent \normalsize Mathias Unberath\footnote{Send correspondence to: \href{mailto:mathias@jhu.edu}{mathias@jhu.edu}.}, Cong Gao, Yicheng Hu, Max Judish,\\
    Russell H Taylor, Mehran Armand, Robert Grupp}
    
    \vspace{10pt}
    
     {\noindent \normalsize Advanced Robotics and Computationally Augmented Environments (ARCADE) Lab\\
     Johns Hopkins University}
     
     \vspace{5pt}\rule{\textwidth}{2pt}
\end{center}

\begin{abstract}
Image-based navigation is widely considered the next frontier of minimally invasive surgery. It is believed that image-based navigation will increase the access to reproducible, safe, and high-precision surgery as it may then be performed at acceptable costs and effort. This is because image-based techniques avoid the need of specialized equipment and seamlessly integrate with contemporary workflows. Furthermore, it is expected that image-based navigation techniques will play a major role in enabling mixed reality environments, as well as autonomous and robot-assisted workflows. A critical component of image guidance is 2D/3D registration, a technique to estimate the spatial relationships between 3D structures, e.\,g., preoperative volumetric imagery or models of surgical instruments, and 2D images thereof, such as intraoperative X-ray fluoroscopy or endoscopy. While image-based 2D/3D registration is a mature technique, its transition from the bench to the bedside has been restrained by well-known challenges, including brittleness with respect to optimization objective, hyperparameter selection, and initialization, difficulties in dealing with inconsistencies or multiple objects, and limited single-view performance. One reason these challenges persist today is that analytical solutions are likely inadequate considering the complexity, variability, and high-dimensionality of generic 2D/3D registration problems. The recent advent of machine learning-based approaches to imaging problems that, rather than specifying the desired functional mapping, approximate it using highly expressive parametric models holds promise for solving some of the notorious challenges in 2D/3D registration. In this manuscript, we review the impact of machine learning on 2D/3D registration to systematically summarize the recent advances made by introduction of this novel technology. Grounded in these insights, we then offer our perspective on the most pressing needs, significant open problems, and possible next steps.

 { \noindent{\bfseries Keywords:} Artificial intelligence, Deep learning, Surgical data science, Image registration, Computer-assisted interventions, Robotic surgery, Augmented reality} 
\end{abstract}

\section{Introduction}

\subsection{Background}

\begin{figure}
    \centering
    \includegraphics[width=\textwidth]{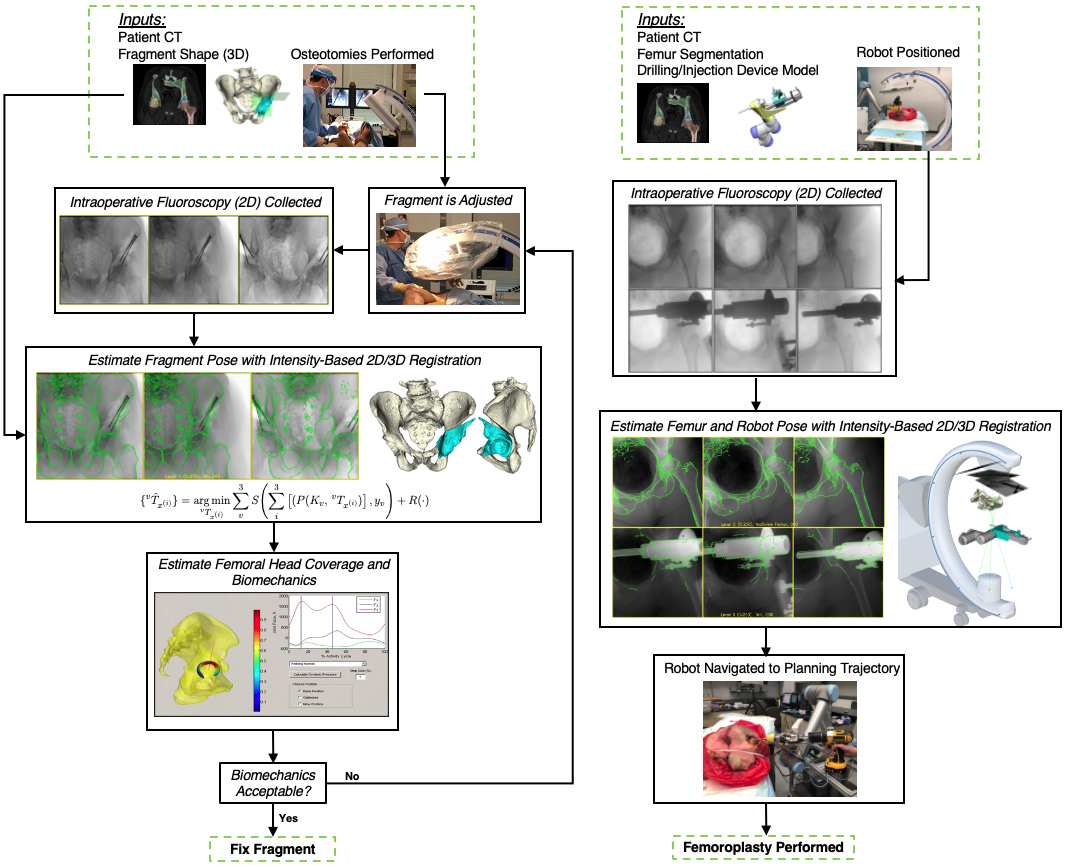}
    \caption{\revised{\textbf{Left:} A high-level overview of the workflow proposed by~\cite{grupp2019pose}, which uses 2D/3D registration for estimating the relative pose of a periacetabular osteotomy (PAO) fragment. By enabling intra-operative 3D visualizations and the calculation of biomechanical measurements, this pose information should allow surgeons to better assess when a PAO fragment requires further adjustments and potentially reduce post-operative complications. The utility of the proposed workflow is diminished by the traditional registration strategy's requirement for manual annotations, which are needed to initialize the pelvis pose and reconstruct the fragment shape.\\
    \textbf{Right:} Image-based navigation for robot-assisted femoropalsty by~\cite{gao2020fiducial_tmrb}. The intra-operative poses of the robot and the femur anatomy are estimated using X-ray-image based 2D/3D registration. The robot-held drilling/injection device is positioned according to the pre-planned trajectory that is propageted intra-operatively using pose estimates from 2D/3D registration. Image-based navigation is less invasive than fiducial-based alternatives and simplifies the procedure.}}
    \label{fig:femoroplasty}
\end{figure}

Advances in interventional imaging, including the miniaturization of high-resolution endoscopes and the increased availability of C-arm X-ray systems, have driven the development and adoption of minimally invasive alternatives to conventional, invasive and open surgical techniques across a wide variety of clinical specialities. While minimally invasive approaches are generally considered safe and effective, the indirect visualization of surgical instruments relative to anatomical structures complicates spatial cognition and the more confined room for maneuvers requires precise command of the surgical instruments. It is well known that due to the aforementioned challenges, among others, outcomes after minimally invasive surgery are positively correlated with technical proficiency, experience, and procedural volume of the operator~\cite{birkmeyer2013surgical,pfandler2019technical,foley2021effectiveness,hafezi2020vertebroplasty}. To mitigate the impact of experience on complication risk and outcomes, surgical navigation solutions that register specialized tools with 3D models of the anatomy using additional tracking hardware are now commercially available~\cite{mezger2013navigation,ewurum2018surgical}. Surgical navigation promotes reproducibly good patient outcomes, and when combined with robotic assistance systems, may enable novel treatment options and improved techniques~\cite{van2016current}. Unfortunately, navigation systems are not widely adopted due to, among other things, high purchase price despite limited versatility, increased procedural time and cost, and potential for disruptions to surgical workflows due to line-of-sight occlusions or other system complications~\cite{picard2014computer,joskowicz2016computer}. While frustrations induced by workflow disruption affect every operator equally, the aforementioned limitations regarding cost particularly inhibit the adoption of surgical navigation systems in geographical areas with less specialized healthcare providers with lower volumes for any procedure; areas where routine use of navigation would perhaps be most impactful.\\ 
To mitigate the challenges of conventional surgical navigation systems that introduce dedicated tracking hardware and instrumentation as well as workflow alterations, the computer-assisted interventions community has contributed purely image-based alternatives to surgical navigation, e.\,g.,~\cite{nolte2000new,mirota2011vision,leonard2016image,tucker2018towards}. Image-based navigation techniques do not require specialized equipment but rely on traditional intra-operative imaging that enabled the minimally invasive technique in the first place. Therefore, these techniques do not introduce economic trade-offs. Further, because image-based navigation techniques are designed to seamlessly integrate into conventional surgical workflows, their use should -- in theory -- not cause frustration or prolonged procedure times~\cite{vercauteren2019cai4cai}. 
A central component to many if not most image-guided navigation solutions is image-based 2D/3D registration, which estimates the spatial relationship between a 3D model of the scene (potentially including anatomy and instrumentation) and 2D interventional images thereof~\cite{markelj2012review,liao2013review}. \revised{Two examples of using 2D/3D registration for intra-operative guidance are shown in Figure~\ref{fig:femoroplasty}: Image-guidance of periacetabular osteotomy (left, and discussed in greater detail in Section~\ref{subsec:multi_object}) and robot-assisted femoroplasty (right).} One may be tempted to assume that after several decades of research on this topic, image-based 2D/3D registration is a largely solved problem. While, indeed, analytical solutions now exist to precisely recover 2D/3D spatial relations under certain conditions~\cite{markelj2012review,uneri20133d,grupp2020fast,gao2020fiducial}, several hard challenges prevail. 

\subsection{Problem formulation}
\label{subsec:problem}

Generally speaking, in 2D/3D registration we are interested in finding the optimal geometric transformation that aligns a (typically pre-operative) 3D representation of objects or anatomy with (typically intra-operative) 2D observations thereof. For the purposes of this review, we will assume that the reduction in dimensionality originates from a projective, not an affine, transformation. \\
Given a set of 3D data~$x_i$ and 2D observations~$y_v$, where subscripts $i,v$ suggest that there may be multiple objects and multiple 2D observations, respectively, a generic way of writing the optimization problem for the common case of a single object but multiple views is:
\begin{equation}
    \{\hat{K}_v, \,^{v}\hat{T}_{x},\hat{\omega}_{D}\} = \underset{K_v, \,^{v}T_{x},\omega_{D}}{\argmin} \sum_v S\!\left( (P(K_v, \,^{v}T_{x}) \circ D_{\omega_{D}})(x), y_v  \right) + R(\cdot)\,.
    \label{eq:reg_objective}
\end{equation}
In Eq.~\ref{eq:reg_objective}, $D_{\omega_{D}}$ is a 3D non-rigid deformation model with parameters $\omega_{D}\in\mathbb{R}^{N_D}$, $P(K_v, \,^{v}T_{x})$ is a rigid projection operation using a camera with intrinsic parameters $K_v\in\mathbb{R}^{3\times3}$ and pose $\,^{v}T_{x} \in SE(3)$ with respect to the 3D data $x$, and $S(\cdot,\cdot)$ is a cost function (in case of images often a similarity metric). We use the composite function $P\circ D(\cdot)$ to capture the variability, including their order, with which these operations may be applied. Finally, $R(\cdot)$ is a regularizing term that can act and/or depend on any combination of variables and parameters; its choice most often depends on the specific application since regularization can represent ``common sense'' or prior knowledge, which tends to vary with the problem domain.   

2D/3D registration then amounts to estimating the  $5+6+N_D$ degrees of freedom (DoFs) for $\{\hat{K}_v, \,^{v}\hat{T}_{x}, \hat{\omega}_{D}\}$, respectively, that minimize the optimization objective. Clearly, there are special cases to Eq.~\ref{eq:reg_objective}, e.\,g., for rigid registration where only $\hat{K}_v, \,^{v}\hat{T}_{x}$ must be estimated and $D_{\omega_{D}}$ is known, or vice versa.\\
In traditional image-based 2D/3D registration, this optimization is usually performed iteratively where parameters are initialized with some values $\{K_v, \,^{v}T_{x}, \omega_{D}\}_0$ and then adjusted incrementally. The updates $\delta \{K_v, \,^{v}T_{x}, \omega_{D}\}$ are derived from Eq.~\ref{eq:reg_objective} using gradient-based or gradient-free methods, such as BFGS~\cite{liu1989limited,berger2016marker} or CMA-ES~\cite{hansen2003reducing} and BOBYQA~\cite{powell2009bobyqa}, respectively. In certain cases when 2D and 3D data representations are not pixel or voxel grid-based but sparse, e.\,g., keypoints, analytic solutions to Eq.~\ref{eq:reg_objective}, such as the perspective n point (PnP) algorithm~\cite{lepetit2009epnp}, exist.

The above traditional approach to solving 2D/3D image registration has spawned solutions that precisely recover the desired geometric transformations under certain conditions~\cite{markelj2012review,uneri20133d,grupp2020fast,gao2020fiducial}. Unfortunately, despite substantial efforts over the past decades, 2D/3D registration is not yet enjoying wide popularity as a workhorse component in image-based navigation platforms at the bedside. Rather, it is shackled to the bench top because several hard open challenges inhibit its widespread adoption. They include:
\begin{itemize}
    \item \textbf{Narrow capture range of similarity metrics}: Conventional intensity-based methods mostly use hand-crafted similarity metrics between $y_v$ and its current best estimate $\hat{y}_v = (P(K_v, \,^{v}T_{x}) \circ D_{\omega_{D}})(x)$ as loss function. Common choices for the similarity metric are Normalized Cross Correlation~(NCC), gradient information~\cite{berger2016marker}, or Mutual Information~(MI)~\cite{maes1997multimodality}. While these metrics are positively correlated with pose differences when the perturbations in $\{\hat{K}_v, \,^{v}\hat{T}_{x},\hat{\omega}_{D}\}$ are small, they are generally non-convex and fail to accurately represent pose offsets when perturbations are large. Thus, without proper initialization, the optimization is prone to get stuck in local minima, returning wrong registration results. The initial estimate of the target parameters~$\{\hat{K}_v, \,^{v}\hat{T}_{x},\hat{\omega}_{D}\}$ must hence be close enough to the true solution in order for the optimization to converge to the global minima. Estimating good initial parameters is commonly achieved using some manual interaction, which is cumbersome and time consuming, or -- in research papers -- neglected all-together. The magnitude by which the initial parameter guesses may be incorrect for the downstream algorithm to still produce a successful registration is referred to as the capture range~\cite{markelj2012review,esteban2019towards}. Its magnitude depends, among other things, on the similarity function as well as the optimizer, and it can be stated quantitatively as the mean target registration error (mTRE) between 3D keypoints at initialization\footnote{It should be noted that the definition of capture range is by no means standardized or even similar across papers, which we will also comment on in Section~\ref{sec:perspective}.}.\\
    The resulting challenges are two-fold: On the one hand, it is important to develop robust and automated initialization strategies for the existing image-based 2D/3D registration algorithms to succeed. On the other hand, there is interest in and opportunity for the development of better similarity metrics that better capture the evolution of image (dis)similarity. Doing so is challenging, however, because of the complexity of the task including various contrast mechanisms, imaging geometries, and inconsistencies.   
    \item \textbf{Ambiguity}: The aforementioned complexity also leads to registration ambiguity, which is most pronounced in single-view registration~\cite{uneri20133d, otake2013robust}. Because the spatial information along the projection line is collapsed onto the imaging plane, it is hard to precisely recover the information in the projective direction. A well-known example is the difficulty of accurately estimating the depth of 3D scene from the camera center using a single 2D image. These challenges already exist for rigid 2D/3D registration and are further exacerbated in rigid plus deformable registration settings. 
    \item \textbf{High dimensional optimization problems}: Even in the simplest case, 2D/3D registration describes a non-convex optimization problem with at least 6 DoFs to describe a rigid body transform. In the context of deformable 2D/3D registration, the high dimensional parameter $\omega_{D}$ that describes the 3D deformation drastically increases the optimization search space. However, since the information within the 2D and 3D images remains constant, the optimization problem may easily become ill-posed. Although statistical modeling techniques exist to limit the parameter search space, the registration accuracy and sensitivity to key features remain an area of concern~\cite{zhu2021iterative}.  
    \item \textbf{Verification and uncertainty}: As a central component of image-based surgical navigation platforms, 2D/3D registration supplies critical information to enable precise manipulation of anatomy. To enable users to assess risk and make better decisions, there is a strong desire for registration algorithms to verify the resulting geometric parameters or supply uncertainty estimates. Perhaps the most straightforward way of verifying a registration result is to visually inspect the 2D overlay of the projected 3D data -- this approach, however, is neither quantitative nor does it scale since it is based on human intervention. 
\end{itemize}
These open problems can be largely attributed to the variability in the problem settings (e.\,g., regarding image appearance and contrast mechanisms, pose variability, \dots) that cannot easily be handled algorithmically because the desired properties cannot be formalized explicitly.
Machine learning methods, including deep neural networks (NNs), have enjoyed a growing popularity across a variety of image analysis problems~\cite{vercauteren2019cai4cai}, precisely because they do not require explicit definitions of complex functional mappings. Rather, they optimize parametric functions, such as convolutional NNs (CNNs), on training data such that the model learns to approximate the desired mapping between input and output variables. 
As such, they provide opportunities to supersede heuristic components of traditional registration pipelines with learning-based alternatives that were optimized for the same task on much larger amounts of data. This allows us to expand Eq.~\ref{eq:reg_objective} into: 
\begin{equation}
    \{\hat{K}_v, \,^{v}\hat{T}_{x},\hat{\omega}_{D}\} = \underset{K_v, \,^{v}T_{x},\omega_{D}}{\argmin} \sum_v S^{\mlemph{\theta_S}}\!\left( (P(K_v, \,^{v}T_{x}) \circ D^{\mlemph{\theta_D}}_{\omega_{D}})(\mlemph{G_x^{\theta_x}(}x\mlemph{)}), \mlemph{G_y^{\theta_y}(}y_v\mlemph{)}  \right) + R^{\mlemph{\theta_R}}(\cdot)\,,
    \label{eq:reg_objective_ML}
\end{equation}
where we have introduced parameters $\theta$ to several components of the objective function to indicate that they may now be machine learning models, such as CNNs. Similarly, the registration may not rely on the original 3D and 2D data itself but some higher-level representation thereof, e.\,g., anatomical landmarks, that are generated using some learned function $G^\theta(\cdot)$. 

In this manuscript, first we summarize a systematic review of the recent literature on machine learning-based techniques for image-based 2D/3D registration, and explain how they relate to Eq.~\ref{eq:reg_objective_ML}. Based on those observations, we identify the impact that the introduction of contemporary machine learning methodology has had on 2D/3D registration for image-guided interventions. Concurrently, we identify open challenges and contribute our perspective on possible solutions.

\section{Systematic review}
\label{sec:review}

\subsection{Search methodology}

The aim of the systematic review is to survey those machine learning-enhanced 2D/3D registration methods in which the 3D data and 2D observations thereof are related through one or multiple perspective projections (and potentially some non-rigid deformation). This scenario arises, for example, in the registration between 3D CT and 2D X-ray, 3D magnetic resonance angiography (MRA) and 2D digital subtraction angiography (DSA), or 3D anatomical models and 2D endoscopy images. 3D/3D registration methods (such as between 3D CT and intra-operative CBCT) or 2D/3D slice-to-volume registration (as it arises, among others, in ultrasound to CT/MR registration) are beyond the scope of this review. 
Because we are primarily interested in surveying the impact of contemporary machine learning techniques, such as deep CNNs, on 2D/3D registration, we limit our analysis to records that appeared after January 2012, which pre-dates the onset of the ongoing surge of interest in learning-based image processing~\cite{krizhevsky2012imagenet}.  

\begin{figure}
    \centering
    \includegraphics[width=0.8\textwidth]{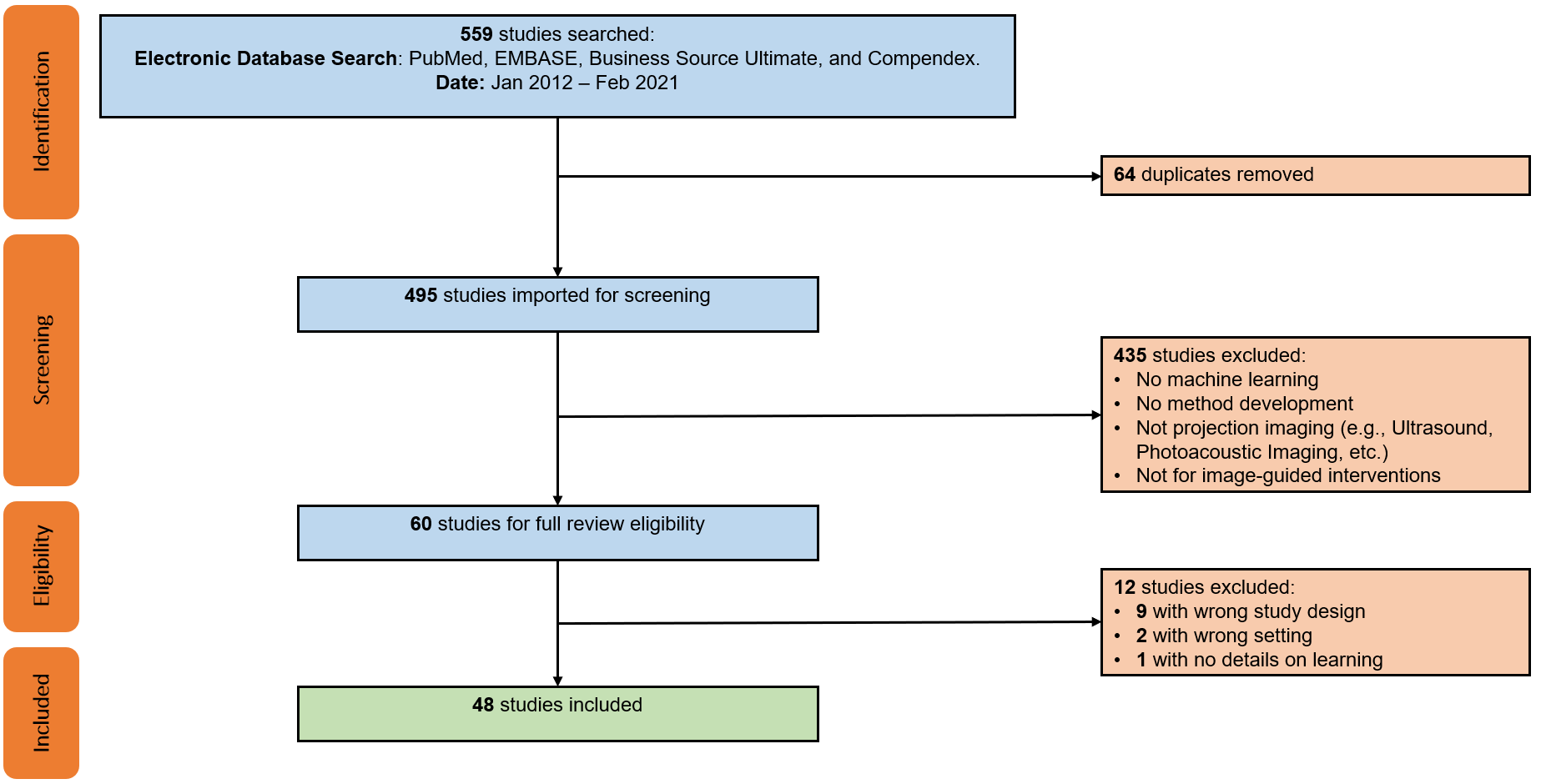}
    \caption{\revised{PRISMA flow chart illustrating the screening and inclusion process. Duplicate studies were the result of searching multiple databases. Exclusion screening was performed individually with each study's abstract with assistance of Covidence tool. Additional twelve studies excluded after full-text review, resulting in a pool of 48 studies included for full review.}}
    \label{fig:PRISMA_chart}
\end{figure}

To this end, we conducted a systematic literature review in accordance with the Preferred Reporting Items for Systematic reviews and Meta-Analyses (PRISMA) method~\cite{moher2009preferred} (cf. Figure~\ref{fig:PRISMA_chart}). We searched PubMed, EMBASE, Business Source Ultimate, and Compendex to find articles pertinent to machine learning for 2D/3D image registration.  The following search terms were used to screen titles, abstracts, and keywords of all available records from January 2012 through February 2021:

(``2D3D registration'' OR ``2D 3D registration'' OR ``3D2D registration'' OR ``3D 2D registration'' OR ``2D/3D registration'' OR ``3D/2D registration'' OR ``2D-3D registration'' OR ``3D-2D registration'' OR ``two-dimensional/three-dimensional registration'' OR ``three-dimensional/two-dimensional registration'') AND (``learning'' OR ``training'' OR ``testing'' OR ``trained'' OR ``tested'')

The initial search resulted in 559 records, and after removal of duplicates, 495 unique studies were included for screening. From those, 447 were excluded because they either did not describe a machine learning-based method for 2D/3D registration, or considered a slice-to-volume registration problem (e.\,g., as in ultrasound or magnetic resonance imaging). The remaining 48 articles were included for an in-depth full text review, analysis, and data extraction, which was performed by five of the authors (M.U., C.G., M.J., Y.H., and R.G.). Initially, every reviewer analyzed 5 articles to develop and refine the data extraction template and coding approach. The final template involved the extraction of the following information: 1) A brief summary of the method including the key contribution; 2) modalities and registration phase (including the 3D modality, 2D modality, the registration goal, whether the method requires manual interactions, whether the method is anatomy or patient-specific, and the clinical speciality), 3) the spatial transformation to be recovered (including the number of objects to be registered, the number of views used for registration, and the transformation model used), 4) information on the machine learning model and training setup (including the explicit machine learning technique, the approach to training data curation as well as to data labeling and supervision, and the application of domain generalization or adaptation techniques), 5) the evaluation strategy (including the data source used for evaluation as well as its annotation, the metrics and techniques used for quantitative and qualitative assessment, and most importantly the deterioration of performance in presence of domain shift), and finally, 6) a more subjective summary of concerns with respect to the experimental or methodological approach or the assumptions made in the design or evaluation of the method. \\
Every one of the 48 articles was analyzed and coded by at least two of the five authors and one author (M.U.) merged the individual reports into a final consensus document.

\subsection{Limitations}

Despite our efforts to broaden the search terms regarding 2D/3D registration, we acknowledge that the list may not be exhaustive. Newer or less popular terminology, such as ``pose regression'', were not used. We also did not use modality specific terms, such as ``video to CT registration'', which may have excluded some manuscripts that focus on endoscopy or other RGB camera-based modalities. The search included terms like ``learning'', ``training'', or ``testing'' as per our interest in machine learning methods for 2D/3D image registration. This search may have excluded some studies that do not explicitly characterize their work as machine or deep ``learning'' and do not describe their training and testing approach, neither in their title nor abstract. \revised{The terminology used in the search may also have resulted in the exclusion of relevant work from the general computer vision literature.} Finally, the review is limited to published manuscripts. Publication bias may have resulted in the exclusion of works relevant to this review.

\subsection{Concise summary of the overall trends}

We first summarize the general application domain and problem setting of the 48 included papers and then review the role that machine learning plays in those applications. The specific characteristics of all included papers are summarized in Tables~\ref{table:registration parameter} and \ref{table:training/testing}, respectively. We state the number of papers either in the running text or in parentheses.

The vast majority of papers (34) considers the 2D/3D registration between X-ray images and CT or cone-beam CT (CBCT) volumes, with the registration of X-ray images and 3D object models being a distant second (10). Other modality combinations included 2D RGB to 3D CT (2) and 3D MR (1), or did not specify the 3D modality (1). The clinical applications that motivate the development of those methods include orthopedics (19), with a focus on pelvis and spine, angiography (9), and radiation therapy (7), e.\,g., for tracking of lung or liver tumors, and cephalometry (4). 
We observe that eleven methods are explicitly concerned with finding a good initial parameter set to begin optimization, while 37 papers (also) describe approaches to achieve high fidelity estimates of the true geometric parameters. Further, four methods consider verification of the registration result. 
Resulting from the clinical task and registration phase distributions, most methods only perform rigid alignment (36), while nine methods consider non-rigid registration only and two approaches address both, rigid and non-rigid registration. To solve the alignment problem, 38 papers relied on a single 2D view, seven approaches used multiple views onto the same constant 3D scene, and two methods assumed the same view but used multiple images of a temporally dynamic 3D scene. 
Perhaps the most striking observation is that all but three (45) included studies only consider the registration of a single object. Two other studies that deal with multiple objects, however, are limited to object detection~\cite{doerr2020data} and inpainting~\cite{esfandiari2021deep}, respectively, and do not report registration results. The remaining study~\cite{grupp2020automatic} performs a 2D segmentation of multiple bones, but does not apply any additional learning to perform the registration.
While there are three methods that do, in fact, describe the 2D/3D registration of multiple objects, i.\,e., vertebral bodies \cite{varnavas2013fully,varnavas2015fullyB} and knee anatomy \cite{wang2020multi}, the individual registrations are solved independently which inhibits information sharing to ease the optimization problem.

\begin{figure}
    \centering
    \includegraphics[width=\textwidth]{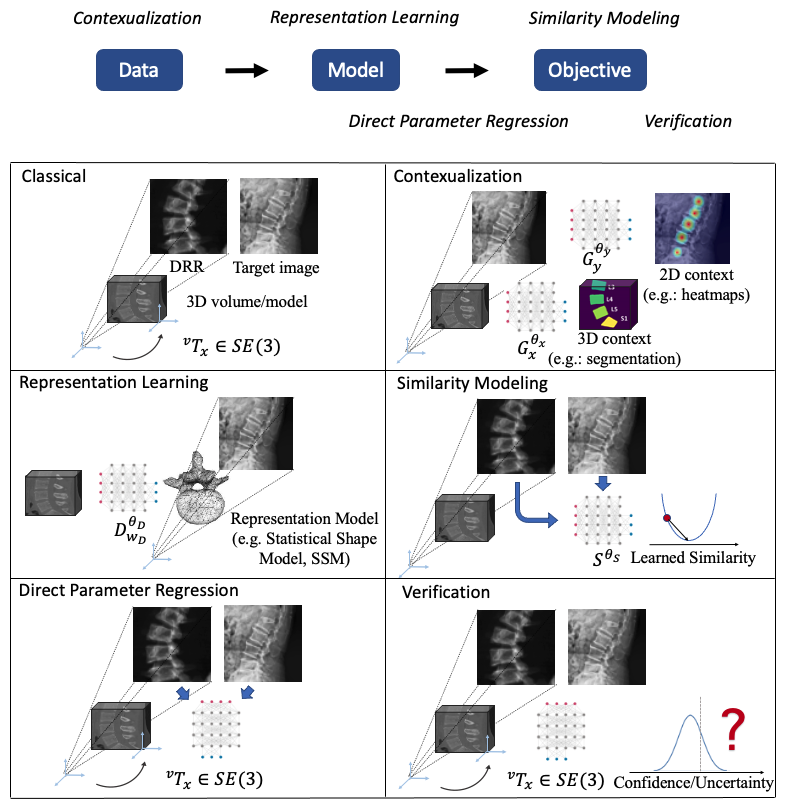}
    \caption{Illustrations of the main themes of machine learning in 2D/3D registration. \revised{The logic relationships of these themes are shown on top.} We use a spine CT volume and a spine X-ray image as an example to show the generic 2D/3D projection geometry. Machine learning models are represented with a neural network icon. Key labels and parameters are presented and map to equations~\ref{eq:reg_objective} and~\ref{eq:reg_objective_ML}.}
    \label{fig:categ}
\end{figure}

The focus of this review is the impact that machine learning has had on the contemporary state of 2D/3D registration and we will briefly introduce the five main themes that we identified here and then discuss them in greater detail in subsequent sections. \revised{From a high-level perspective of abstracting 2D/3D registration problems, they follow the flow of acquiring \textit{Data}, fitting \textit{Model} and solving the \textit{Objective}. The five themes which we categorize logically aim at improving certain aspects of this flow.} The themes are: 
\begin{itemize}
    \item Contextualization (Section~\ref{subsec:context}): Instead of relying solely on the images themselves, the 14 methods in this theme use machine learning algorithms to extract semantic information from the 2D or 3D data, including landmark or object detection, semantic segmentation, or data quality classification. Doing so enables automatic initialization techniques, sophisticated regularizers, as well as techniques that handle inconsistencies between 2D and 3D data.
    \item Representation learning (Section~\ref{subsec:repre}): Principal component analysis (PCA), among other techniques, are a common way to reduce the dimensionality of highly complex data -- in this case, rigid and non-rigid geometric transformations. Twelve papers used representation learning techniques as part of the registration pipeline. 
    \item Similarity modeling (Section~\ref{subsec:repre}): Optimization-based image registration techniques conventionally rely on image similarity metrics that ideally should capture appearance differences due to both large and very fine scale geometric misalignment. Ten studies describe learning-based approaches to improve on similarity quantification.  
    \item Direct parameter regression (Section~\ref{subsec:regress}): In contrast to iterative methods, direct parameter regression techniques seek to infer the correct geometric parameters for 2D/3D alignment (either absolute with respect to a canonical 3D coordinate frame, or relative between a source and a target coordinate frame) directly from the 2D observation. A total of 22 manuscripts reported such approaches for either rigid or non-rigid registration.    
    \item Verification (Section~\ref{subsec:veri}): Four studies used machine learning-based techniques to assess whether the estimated geometric parameters should be considered reliable. 
\end{itemize}
High-level depictions of these themes are shown in figure~\ref{fig:categ} and the respective sections below provide details for each, along with references to the individual studies.

\subsection{Contextualization}
\label{subsec:context} 
Studies summarized in this theme use machine learning techniques to increase the information available to the 2D/3D registration problem by extracting semantic information from the 2D or 3D data~\mlemph{\cite{lin2012shape,francoise2020detecting,esfandiari2021deep,grupp2020automatic,bier2019learning,karner2020single,luo2019towards,varnavas2013fully,varnavas2015fullyB,doerr2020data,wang2020multi,yang2019a,bier2018x,chen2018real}}.\\
Using the notation of Eq.~\ref{eq:reg_objective_ML}, these methods specify $G_y^{\theta_y}(y), G_x^{\theta_x}(x)$, or $R^{\theta_R}(\cdot)$ although not all methods are necessarily integrated in the iterative optimization procedure. Perhaps the most prevalent approach here is the detection of anatomical landmarks on 2D images~\cite{grupp2020automatic,bier2019learning,bier2018x,wang2020multi,yang2019a} to define correspondences with the respective 3D locations, which allows for either, explicit determination of $\hat{y}_v =\{K_v, \,^{v}T_{x}\}$ using PnP~\cite{bier2019learning,bier2018x} or stereo-reconstruction following 3D-3D matching~\cite{wang2020multi,yang2019a}, or for the introduction of soft re-projection constraints as a regularizing term~\cite{grupp2020automatic}. Another way of benefiting initialization through contextualization is object detection~\cite{lin2012shape,varnavas2013fully,varnavas2015fullyB,doerr2020data,chen2018real}. While all methods for landmark detection rely on deep CNNs, object detection achieved satisfactory results already using less complex learning models, i.\,e., templates in~\cite{lin2012shape}, PCA over object contours in~\cite{chen2018real}, and the Generalized Hough Transform in \cite{varnavas2013fully,varnavas2015fullyB}. The drawback, however, is that most of these less complex approaches require patient-specific training since the models are unable to generalize beyond a single shape. \cite{doerr2020data} describe a deep learning-based alternative, where the Fast R-CNN object detector is re-trained to return bounding boxes of 30 different screw types in varied poses.\\
A complementary trend is the identification of image regions~\cite{francoise2020detecting,esfandiari2021deep} or whole images~\cite{luo2019towards} that should not contribute to the optimization problem because of inconsistency. \cite{francoise2020detecting} use a U-net-like fully convolutional NN (FCN) to segment occluding contours of the uterus to reject regions while \cite{luo2019towards} identify and reject poor quality frames in bronchoscopy. \cite{esfandiari2021deep} consider mismatch introduced by intra-operative instrumentation. They contribute a non-blind image inpainting method using a FCN that seeks to restore the background anatomy after image regions corresponding to instruments were identified.\\
It is undeniable that the introduction of machine learning to contextualize 2D and 3D data enables novel techniques that quite substantially expand the tools one may rely on when designing a 2D/3D registration algorithm, and as such, are likely to become impactful. However, a general trend that we observed in most of these studies was that the impact of the contextualization component on the downstream registration task was not, in fact, evaluated. For example, while \cite{bier2019learning} report quantitative results on real data of cadaveric specimens, it remains unclear whether the performance would be sufficient to actually initialize an image similarity-based 2D/3D registration algorithm. There are, of course, positive examples including~\cite{varnavas2015fullyB,luo2019towards,grupp2020automatic} that demonstrate the benefit of contextualization on overall pipeline performance -- Empirical demonstrations should strongly be preferred over arguments from authority.

\subsection{Representation learning}
\label{subsec:repre} 
As highlighted in Section~\ref{subsec:problem}, 2D/3D registration, especially in deformable scenarios, suffers from high dimensional parameter spaces and changes in any one of those parameters are not easily resolved due to limited information which creates ambiguity. In our review we found that unsupervised representation learning techniques are a widely adopted technique to reduce the dimensionality of the parameter space while introducing implicit regularization by confining possible solutions to the principal modes of variation across population- or patient-level observations. We identified 12 studies that propose such techniques or use them as part of the registration pipeline    \mlemph{\cite{brost2012constrained,zhang2020automatic,zhao2014local,baka2015respiratory,chou20132d,chou2014local,chou2013real,chen2018real,li2020non,foote2019real,zhang2018temporal,pei2017non,lin2012shape}}.\\ 
PCA is by far the most prevalent method for representation learning and is used in all but one study. This specific study, however, used by far the most views $v=20$ for initial estimation of a low resolution vector field, which was then regularized by projection onto a deep learning-based population model~\cite{zhang2020automatic}. We found that methods designed for cephalometry were distinct from all other approaches as their primary goal is not generally 2D/3D registration, but 3D reconstruction of the skull given a 2D X-ray. Among the papers included in this review, this problem is often formulated as the deformable 2D/3D registration between a lateral X-ray image of the skull and a 3D atlas using a PCA deformation model, the principal components $\omega_{D}$ of which are estimated via a prior set of 3D/3D registrations~\cite{pei2017non,zhang2018temporal,li2020non}. Consequently, these methods rely on population-level models and are thus different from methods used for radiation therapy~\cite{zhao2014local,chou20132d,chou2013real,chou2014local,foote2019real} and angiography~\cite{brost2012constrained,baka2015respiratory}, which rely on patient-specific models that are built pre- and intra-operatively, respectively. It is worth mentioning that, while most methods rely on PCA to condense deformable motion parametrizations, it has also been found useful to identify and focus on the primary modes of variation in rigid registration~\cite{brost2012constrained}.\\
We note that most studies in this theme do not consider rigid alignment prior to deformable parameter estimation, e.\,g., by assuming perfectly lateral radiographs of the skull for cephalometry or perfectly known imaging geometry in radiation therapy. Except for few exceptional cases with highly specialized instrumentation, for example~\cite{chou2013real} that relied on on-board CBCT imaging, assumptions around rigid alignment seem to be unjustified, which may suggest that the performance estimates need to be interpreted with care. This is further emphasized by the fact that many studies are only evaluated on synthetic data (which we fear may have sometimes been generated by sampling the PCA model also used for registration, introducing inverse crime) and may not have paired 3D data for extensive quantitative evaluation.

\subsection{Similarity modeling}
\label{subsec:sim}

As we established in Section~\ref{subsec:problem} for optimization-based 2D/3D registration algorithms, the cost function -- or similarity metric -- $S(\cdot,\cdot)$ is among the most important components since it will determine parameter updates. It is well known that most commonly used metrics fail to accurately represent the distances in geometric parameter space that generate the mismatch between the current observations. It is thus not surprising that ten studies describe methods to better model and quantify the similarity between the source and target images to increase the capture range, and thus, the likelihood of registration success \mlemph{\cite{francoise2020detecting,schaffert2020learning,tang2016similarity,schaffert2019metric,schaffert2020learningB,neumann2020deep,gu2020extended,liao2019multiview,grupp2020automatic,gao2020generalizing}}.\\
Some studies propose novel image similarity functions $S^{\theta_S}(\cdot,\cdot)$ that, analogous to traditional similarity metrics, accept as input the source and target image and return a scalar or vector that is related to the mismatch in parameter space~\cite{francoise2020detecting,gu2020extended,tang2016similarity,neumann2020deep,grupp2020automatic,gao2020fiducial}. Among those, two methods rely on regularization: \cite{grupp2020automatic} detect anatomical landmarks to expand an analytic similarity function with landmark-reprojection constraints to enhance the capture range of an intensity-based strategy, while \cite{francoise2020detecting} segment occluding contours to constrain similarity evaluation to salient regions. The other four methods use machine learning models to approximate a geometric parameter distance function based on the input images. To this end, \cite{gu2020extended} and \cite{gao2020fiducial} estimate the geodesic in Riemannian tangent space between the source and target camera poses, which in an ideal case results in a convex similarity function. \cite{tang2016similarity} learn a more expressive feature descriptor to better quantify the mismatch in vasculature registration, and for a similar application, \cite{neumann2020deep} regress the disparity between corresponding points in source and target images to quantify image dissimilarity. Both methods rely on concepts that are limited to sparse objects, such as vessels. 
Different to image-based similarity metrics, four studies describe methods for keypoint matching to compute image similarity~\cite{schaffert2019metric,schaffert2020learning,schaffert2020learningB,liao2019multiview}. To this end, \cite{liao2019multiview} train a network to establish keypoint correspondences between the source and target images. Because the geometric parameters of the source image are known, the unknown target parameters can be recovered relative to the source image using PnP-like methods. Finally, a series of three papers~\cite{schaffert2019metric,schaffert2020learning,schaffert2020learningB} characterizes a learning-based method to adaptively weight point correspondences that are established to quantify the degree of misalignment.\\
As methods in this theme have primarily focused on expanding the capture range of contemporary similarity metrics, the potential shortcomings of other components of the registration pipeline, such as the optimizers, remain unaffected. While we introduced \cite{gao2020fiducial} in the context of similarity learning, the method also describes a fully differentiable 2D/3D registration pipeline that addresses optimization aspects. This enables both, end-to-end learning of $G_y^{\theta_y}(y), G_x^{\theta_x}(x)$ and/or $S^{\theta_S}(\cdot,\cdot)$ during training as well as analytic gradient-based optimization using backpropagation during application. Further, we found that most emphasis was given on increasing the capture range of the registration pipeline and very little, if any, attention is paid to increasing the resolution and precision of these metrics. Especially in single-view registration scenarios, which we have identified to be most prevalent, it is well known that certain DoFs cannot be resolved with high accuracy. Therefore, developing methods that not only increase the capture range but also the precision of 2D/3D registration pipelines should be of high priority.

 \subsection{Direct parameter regression}
\label{subsec:regress} 

So far and especially in the context of Section~\ref{subsec:sim}, 2D/3D registration was motivated as an optimization-based process that compares the source image, generated using the current geometric parameter estimate, with the desired target $y_v$ using some cost function. However, this problem can also be formulated in the context of regression learning, where a machine learning algorithm directly predicts the desired geometric parameters $K_v, \,^{v}T_{x},$ and/or $\omega_{D}$ from $y_v$, or from both $y_v$ and $\hat{y}_v$. Such methods partially or completely skip the step of precise modeling of image formation or similarity, and instead build up the knowledge in a data-driven manner. We have identified 22 studies that describe methods for direct parameter regression \mlemph{\cite{xiangqian20202d,wu2015fully,toth20183d,hou2017predicting,miao2016a,zhao2014local,chou20132d,chou2014local,chou2013real,xie2017single,hou2018computing,guan2020transfer,guan2019transfer,li2020non,foote2019real,zhang2018temporal,pei2017non,miao2016real,gao2020generalizing,miao2018dilated,mitrovi2015simultaneous,zheng2018pairwise}}.\\
Relying on parameter regression solely based on the target image $y_v$ is particularly prevalent for radiation therapy, where the main application is the regression of the principal components of a patient-specific PCA motion model~\cite{zhao2014local,chou20132d,chou2014local,chou2013real,foote2019real}. The importance of regression learning is primarily attributed to the substantially decreased run-time that enables close to real-time tumor tracking in 3D. Methods directed at cephalometry~\cite{li2020non,zhang2018temporal,pei2017non} are identical in methodology to the radiation therapy methods. As noted in Section~\ref{subsec:repre}, most methods here limit themselves to shape estimation and assume that a global rigid alignment is either performed prior or unnecessary. 
The remaining 14 methods consider rigid parameter regression, and we differentiate methods that infer pose directly from the target $y_v$~\cite{xiangqian20202d,wu2015fully,hou2017predicting,hou2018computing,xie2017single,guan2019transfer,guan2020transfer}, and methods that process both $y_v$ and $\hat{y}_v$~\cite{toth20183d,miao2016a,miao2016real,miao2018dilated,zheng2018pairwise,gao2020generalizing,mitrovi2015simultaneous}, and therefore, can run iteratively. 
Methods that rely on the target image only are relatively straight-forward and generally train a standard feed-forward CNN architecture to regress pose on large datasets comprising of multiple independent objects or anatomies. While the simplicity of these approaches is appealing, a general concern is that poses are absolute which requires the definition of a canonical 3D coordinate system. This challenge is mitigated for applications that consider an instrument or tool, because such a canonical system can be readily defined; however, establishing this reference frame is considerably more effortful for patient anatomy and may require (group-wise) 3D/3D registrations. Unfortunately, none of the studies reviewed here describes a dedicated effort to establish such a canonical reference frame, suggesting that those methods will eventually hit a performance ceiling that is determined by the misalignment within the reference coordinate systems in the training data. 
Methods that regress pose between $y_v$ and $\hat{y}_v$ avoid the aforementioned concern because poses are relative rather than absolute, however similarly to conventional techniques, they require an initialization. There is some flexibility in how the information from source and target images is combined. \cite{miao2016a,miao2016real,miao2018dilated,zheng2018pairwise} use the image residual between $y_v$ and $\hat{y}_v$ in a region of interest around the projected tool model. Because only a small part of the target image is used, the initialization must be sufficiently close. \cite{toth20183d} extract features using a CNN from both source and target image independently and then concatenate them for pose regression using fully connected layers. Rather than regressing pose directly, \cite{gao2020generalizing} introduce a fully differentiable 2D/3D pipeline and compare $G_y^{\theta_y}(y)$ and $G_y^{\theta_y}(\hat{y})$ using a simple $L_2$ distance as the similarity function $S(\cdot,\cdot)$. The parameters $\theta_y$ and $\theta_x$ are then optimized using a double backward pass on the computational graph, such that the gradient $\nicefrac{\partial S}{\partial \,^{v}T_x}$ aligns with the geodesic between the current pose estimate and the desired target pose.\\
Pose regression directly from images is appealing because it may result in substantially faster convergence, potentially with a single forward pass of a CNN. We found that all methods included in this review are limited to registration of a single object and it remains unclear how these methods would apply to multiple objects, in part, because of the combinatorial explosion of relative poses. Further, methods that solely rely on the target image never involve neither the 3D data nor the source images created from it. This may be problematic, because it is unclear how these methods would verify that this specific 2D/3D registration data satisfy, among other things, the canonical coordinate frame assumption.

\subsection{Verification}
\label{subsec:veri}
We identified four studies that leverage machine learning techniques for verifying whether a registration process produced satisfactory geometric parameter estimates~\mlemph{\cite{wu2016a,mitrovi2014automatic,varnavas2013fully,varnavas2015fullyB}}.\\ 
An interesting observation is that none of the included methods make use of the resulting images or overlays, but rather rely on low-dimensional data. \cite{varnavas2013fully,varnavas2015fullyB}, for example, rely on the cost function value and the relative poses of multiple objects that are registered independently as input to a support vector machine classifier. Similarly, \cite{wu2016a} train a shallow NN to classify registration success based on hand-crafted features of the objective function surface around the registration estimate. Finally, \cite{mitrovi2014automatic} compare a registration estimate to known local minima and thresholds to determine success/failure, which worked well but may be limited in practice as the approach seems to assume knowledge of the correct solution.\\ 
Some studies included in this theme stand out, in that they are certainly mature and were demonstrated to work well on comparably large amounts of real clinical data, such as \cite{varnavas2015fullyB} and \cite{wu2016a}; unfortunately however, these methods are not general purpose as they rely on the registration of multiple objects and the availability of two orthogonal views, respectively. Compared to the other four themes and maybe even in general, there has been very little emphasis on and innovation in the development of more robust and general purpose methods for the verification of 2D/3D registration results, which we perceive to be a regrettable omission.

\section{Perspective}
\label{sec:perspective}

The introduction of machine learning methodology to the 2D/3D registration workflow was partly motivated by persistent challenges, which were not yet satisfactorily addressed by heuristic and purely algorithmic approaches. Upon review of the recent literature in Section~\ref{sec:review}, perhaps the most pressing question is: Has machine learning resolved any of those open problems? We begin our discussion using the categorization of Section~\ref{subsec:problem}: 
\begin{itemize}
    \item \textbf{Narrow capture range of similarity metrics}: We identified many methods that quantitatively demonstrate increased capture range of 2D/3D registration pipelines. The means to accomplishing this, however, are diverse. Several methods describe innovative learning-based similarity metrics that better reflect distances in geometric parameter space, while other studies present (semi-)global initialization or regularization techniques which may rely on contextual data. The demonstrated improvements are generally significant, suggesting huge potential for machine learning in regards to this particular challenge. Contextualization, such as landmark detection and segmentation, can mimic user input to enable novel paradigms for initialization while also providing a clear interface for human-computer interaction~\cite{amrehn2017ui,zapaishchykova2021,du2019techniques}. Similarity modeling using learning-based techniques -- potentially combined with contextual information -- is a similarly powerful concept that finds broad application also in slice-to-volume registration, e.\,g., for ultrasound to MRI~\cite{hu2018label}, which was beyond this review. 
    \item \textbf{Ambiguity}: While several methods report an overall improved registration performance when using their novel, machine learning-enhanced algorithms, we did not register any method with a particular focus on reducing ambiguity. Because performance is usually reported as a summary statistic over multiple DoFs and instances, it is unclear to what extent performance increases should be attributed to 1) registering individual instances more precisely (which would suggest reduced ambiguity) or 2) succeeding more often (which would rather emphasize the importance of the capture range).  
    \item \textbf{High dimensional optimization problems}: We found that representation learning techniques, currently dominated by PCA, are a clearly established tool to reduce the dimensionality of deformable 2D/3D registration problems, and may even be useful for rigid alignment. In those lower dimensional spaces, e.\,g., the six parameters of a rigid transformations or the principal components of a PCA model, direct pose regression from the target and/or source images is a clearly established line of research. These approaches supersede optimization with a few forward passes of a machine learning model, which makes them comparably fast. This is particularly appealing for traditionally time critical applications, such as tumor tracking in radiation therapy. Complementary approaches that seek to enable fully differentiable 2D/3D registration pipelines for end-to-end training and analytical optimization of geometric parameters, such as~\cite{gao2020generalizing,shetty2021deep}, combine elements of optimization and inference. While these methods are in a development stage, their flexibility may prove a great strength in developing solutions that meet clinical needs.\\ 
    Certainly, some of the above ideas will become increasingly important in the quest to accelerate and improve 2D/3D registration pipelines. This is because image-based navigation techniques~\cite{sugano2003computer,tucker2018towards,hummel20082d} as well as visual servoing of surgical robots~\cite{gao2019localizing,unberath2019enabling,yi2018robotic} will also require high-precision 2D/3D registration at video frame-rates.  
    \item \textbf{Verification and uncertainty}: Compared to the other challenges and themes, very few studies used machine learning to benefit verification of registration results. The studies that did, however, reported promising performance even with rather simple machine learning techniques on low dimensional data, i.\,e., cost function properties rather than images themselves. Quantifying uncertainty in registration slowly emerges as a research thrust in 2D/2D and 3D/3D registration~\cite{pluim2016truth,sinha2019endoscopic}. It is our firm belief that the first generally applicable methods for confidence assignment and uncertainty estimation in 2D/3D registration will become trend-setting due to the nature of the clinical applications that 2D/3D registration enables.  
\end{itemize}
Despite this positive prospect on the utility of machine learning for 2D/3D registration, we noted certain trends and recurring shortcomings in our review that we will discuss next. As was done before, we either specify the number of studies satisfying a specific condition in the text or state it in parentheses.

\subsection{Preserving improvements under domain shift from training to deployment}
\label{subsec:generalization}
An omnipresent concern in the development of machine learning-based components for 2D/3D registration, but everywhere really, is the availability of or access to large amounts of relevant data. In some cases, the data problem amounts to a simple opportunity cost, e.\,g., for automation of manually performed tasks such as landmark detection. It should be noted, however, that even for this ``simple'' case to succeed, many conditions must be met including  ethical review board approval, digital medicine infrastructure, and methods to reliably annotate the data. In many other -- from a research perspective perhaps more exciting -- cases, this retrospective data collection paradigm is infeasible because the task to be performed with a machine learning algorithm is not currently performed in clinical practice. The more obvious examples are visual servoing of novel robotic surgery platforms~\cite{gao2019localizing} or robotic imaging paradigms that alter how data is acquired~\cite{zaech2019learning,thies2020learning}. 
Despite the fact that most studies included in this review address use-cases that fall under the ``opportunity cost'' category, we found that only 16 out of the 48 studies used real clinical or cadaveric data to train the machine learning algorithms. All remaining papers relied on synthetic data, namely digitally reconstructed radiographs (DRRs), that were simulated from 3D CT scans to either replace or supplement (small) real datasets.\\
Training on synthetic data has clear advantages because large datasets and corresponding annotations can be generated with relatively little effort. In addition, rigid pose and deformation parameters are perfectly known by design thus creating an unbiased learning target. Contemporary deep learning-based techniques enable the mapping of very complex functions directly from high-dimensional input data at the cost of heavily over-parameterized models that require as much data as possible to learn sensible associations. These unrelenting requirements, especially with respect to annotation, are not easily met with clinical data collection. Indeed, of the 32 studies that describe deep learning-based methods, 24 trained on synthetic data (seven trained on real data, and one did not train at all but used a pre-trained network). It is evident that data synthesis is an important idea that enables research on creative approaches that contribute to the advancement of 2D/3D registration.\\
Unfortunately, there are also substantial drawbacks of synthetic data training. Trained machine learning algorithms approximate the target function only on a compact domain~\cite{zhou2020universality}, and their behaviour outside this domain is unspecified. Because synthesized data is unlikely to capture all characteristics of data acquired using real systems and from real patients, the domains defined by the synthetic data used for training and the real data used during application will not, or only partially, overlap. This phenomenon is known as domain shift. Therefore, applying a synthetic data-trained machine learning model to real data is likely to result in substantially deteriorated performance~\cite{unberath2018deepdrr,unberath2019enabling}. While this problem exists for all machine learning algorithms, it is particularly prevalent for modern deep learning algorithms as they operate on high-resolution images directly, where mismatch in characteristics (such as noise, contrast, \dots) is most pronounced.\footnote{Non-deep learning techniques usually operate on lower dimensional data that is abstracted from the images (such as cost function values~\cite{wu2016a} or centerlines~\cite{tang2016similarity}) such that domain shift is handled elsewhere in the pipeline, e.\.g., in a segmentation algorithm.} 
Indeed, among the seven studies that trained deep CNNs on synthetic data and evaluated in a way that allowed for comparisons between synthetic and real data performance, we found quite substantial performance drops~\cite{miao2018dilated,li2020non,gao2020generalizing,gu2020extended,bier2019learning,doerr2020data,guan2020transfer}. Worse, three studies used different evaluation metrics in synthetic and real experiments so that comparison was not possible~\cite{toth20183d,esfandiari2021deep,miao2016a}, and perhaps worst, ten studies that trained on synthetic data never even tested (meaningfully) on real data~\cite{zhang2020automatic,hou2017predicting,hou2018computing,xie2017single,neumann2020deep,yang2019a,bier2018x,guan2019transfer,foote2019real,pei2017non}. To mitigate the negative impact of domain shift, there is growing interest in domain adaptation and generalization techniques. Several methods do, in fact, already incorporate some of those techniques (\cite{zheng2018pairwise,wang2020multi} use domain adaptation to align feature representations of real and synthetic data: \cite{gu2020extended,esfandiari2021deep,foote2019real} use heavy pixel-level transformations that approximate domain randomization, and \cite{xiangqian20202d,bier2018x,bier2019learning,gu2020extended,gao2020generalizing,miao2016a} rely on realistic synthesis to reduce domain shift, e.\.g., using open-source physics-based DRR engines~\cite{unberath2018deepdrr,unberath2019enabling}). However, as we have outlined above their impact is not yet strongly felt. For the new and exciting 2D/3D registration pipelines reviewed here to impact image-based navigation, we must develop novel techniques to increase the robustness under domain shift to preserve the method's level of performance when transferring from training to deployment domain.

\subsection{Experimental design, reporting, and reproducibility}
Quantitatively evaluating registration performance is clearly important. The error metrics that were used included the standard registration pose error (commonly separated into translational and rotational DoFs), keypoint distances~\cite{chou2014local, zhang2020automatic}, and the mean target registration error (mTRE) in 3D, and varied other metrics in 2D, such as reprojection distances~\cite{bier2019learning}, segmentation or overlap DICE score~\cite{zhang2020automatic}, or contour differences~\cite{chen2018real}. Other metrics that are not uniquely attributable to a domain include the registration capture range~\cite{schaffert2019metric, esfandiari2021deep} and the registration success rate~\cite{varnavas2013fully, mitrovi2014automatic, miao2016real, yang2019a, schaffert2020learning}. There are no standard routines to define the successful registrations.\\
The wealth of evaluation strategies and metrics can, in part, be attributed to the fact that different clinical applications necessitate different conditions to be met. For example, in 2D/3D deformable registration for tumor tracking during radiation therapy, accurately recovering the 3D tumor shape and position (quantified well using, e.\,g., the DICE score of true and estimated 3D position over time) is much more relevant than a Euclidean distance between the deformation field parameters, which would describe irrelevant errors far from the region of interest. We very clearly advocate \emph{for} the use of task-specific evaluation metrics, since ultimately those metrics are the ones that will distinguish success from failure in the specific clinical application. However, we also believe that the lack of universally accepted reporting guidelines, error metrics, and datasets is a severe shortcoming that has unfortunate consequences, such as a high risk of duplicated efforts and non-interpretable performance reporting. We understand the most pressing needs to be: 
\begin{itemize}
    \item \textbf{Standardizing evaluation metrics}: An issue that appears to have become more prevalent with the introduction of machine learning methods that are developed and trained on synthetic data is the lack of substantial results on clinically relevant, real data. While experimental conditions, including ground truth targets, are perfectly known for simulation, they are much harder to obtain for clinical or cadaveric data. A common approach to dealing with this situation is to provide detailed quantification of mTRE, registration accuracy, etc. on synthesized data where the algorithm will perform well (cf. Section~\ref{subsec:generalization}) while only providing much simpler, less informative, and sometimes purely qualitative metrics for real data experiments. Clearly, this practice is undesirable because 1) synthetic data experiments now cannot serve as a baseline (since they use different metrics and are thus incomparable), and 2) the true quantities of interest remain unknown (for example, a 3D mTRE is more informative than a 2D reprojection TRE since it can adequately resolve depth).\\
    While it is evident that not all evaluation paradigms that are easily available on synthetic data can be readily transferred to clinical data, the reverse is not true. If real data experiments require simplified evaluation protocols because some gold standard quantities cannot be assessed, then these simplified approaches should at a minimum also be implemented on synthetic data to further complement the evaluation. While this approach may still leave some questions regarding real data performance unanswered, it will at least provide reliable information to assess the deterioration from sandbox to real life. 
    \item \textbf{Reporting problem difficulty}: A confounding factor that needs to be considered even when consistent metrics are being used is the fact that different datasets are likely to posit 2D/3D registration problems of varied difficulty. For example, evaluation on synthetic data may include data sampled from a broad range of viewpoints or deformations that are approximately uniformly distributed. Real data, on the other hand, is not uniformly sampled from such a distribution, but rather, will be clustered around certain viewpoints (see \cite{grupp2020automatic} for a visualization of viewpoints used during a cadaveric surgery vs. the synthetic data for the same machine learning task in \cite{bier2019learning}). Because those viewpoints are optimized for human interpretation, they are likely to contain more easily interpretable information which suggests that algorithmic assessment is also more likely to succeed. Then, in those cases, even if the quantitative metrics reported across those two datasets would suggest similar performance, in reality there is degradation due to evaluation on a simpler problem.\\ 
    One way to address this challenge would be to attempt the harmonization of problem complexity by recreating the real dataset synthetically. Another, perhaps more feasible approach would be to develop reporting guidelines that allow for a more precise quantification of problem complexity, e.\,g., by more carefully describing the variation in the respective datasets.  
    \item \textbf{Enabling reproduction}: The recent interest in deep learning has brought about a transformational move towards open science, where large parts of the community share their source code publicly and demonstrate the performance of their solutions on public benchmarks, which allows for fair comparisons. While a too strong focus on ''winning benchmarks'' is certainly detrimental to the creativity of novel approaches, we feel that the lack of public benchmarks for 2D/3D registration and related problems is perhaps an even greater worry. In addition to the use of different metrics for validation purposes discussed above, studies may even use different definitions of the same quantity (such as the capture range, that is defined, e.\,g., using the decision boundary of an SVM classifier in~\cite{esfandiari2021deep} and using mTRE in~\cite{schaffert2020learning}). Further, most code-bases and more importantly datasets are kept private which inhibits reproduction, since re-implementation is particularly prone to biased conclusions. Consequently, creating public datasets with well-defined and standardized testing conditions should be a continued and reinforced effort, and wherever possible, the release of source code should be considered.\\
    To this end, our group has previously released a relatively large dataset of CTs and $>350$ X-rays across multiple viewpoints of 6 cadaveric specimens prior to undergoing periacetabular osteotomy (cf. Fig.~\ref{fig:pao_chisel})~\cite{T1/IFSXNV_2020}. Further, we have made available the core registration components of our intensity-based framework, \href{https://github.com/rg2/xreg}{xReg}, as well as our open-source framework for fast and physics-based synthesis of DRRs from CT, \href{https://github.com/arcadelab/deepdrr}{DeepDRR}, which may allow for the creation of semi-realistic but very large and precisely controlled data. Increasing the rate with which we share code and data will likely result in research that truly advances the contemporary capabilities since the baselines are transparent and much more clearly defined. 
\end{itemize}
These challenges clearly restrict research, but even more dramatically inhibit translational efforts, simply because we cannot reliably understand whether a specific 2D/3D registration problem should be considered ``solved'' or what the open problems are. Finding answers to the posited questions will become especially important as 2D/3D registration technology matures and is integrated in image-based navigation solutions that are subject to regulatory approval. Then, compelling evidence will need to be provided on potential sources and extent of algorithmic bias as well as reliable estimates of real world performance~\cite{us2021artificial}. Adopting good habits around standardized reporting will certainly be a good first step to translate research successes into patient outcomes through productization.

\subsection{Registration of multiple objects, compound or non-rigid motion, and presence of foreign objects}
\label{subsec:multi_object}%
\begin{figure}
    \centering
    \includegraphics[width=\textwidth]{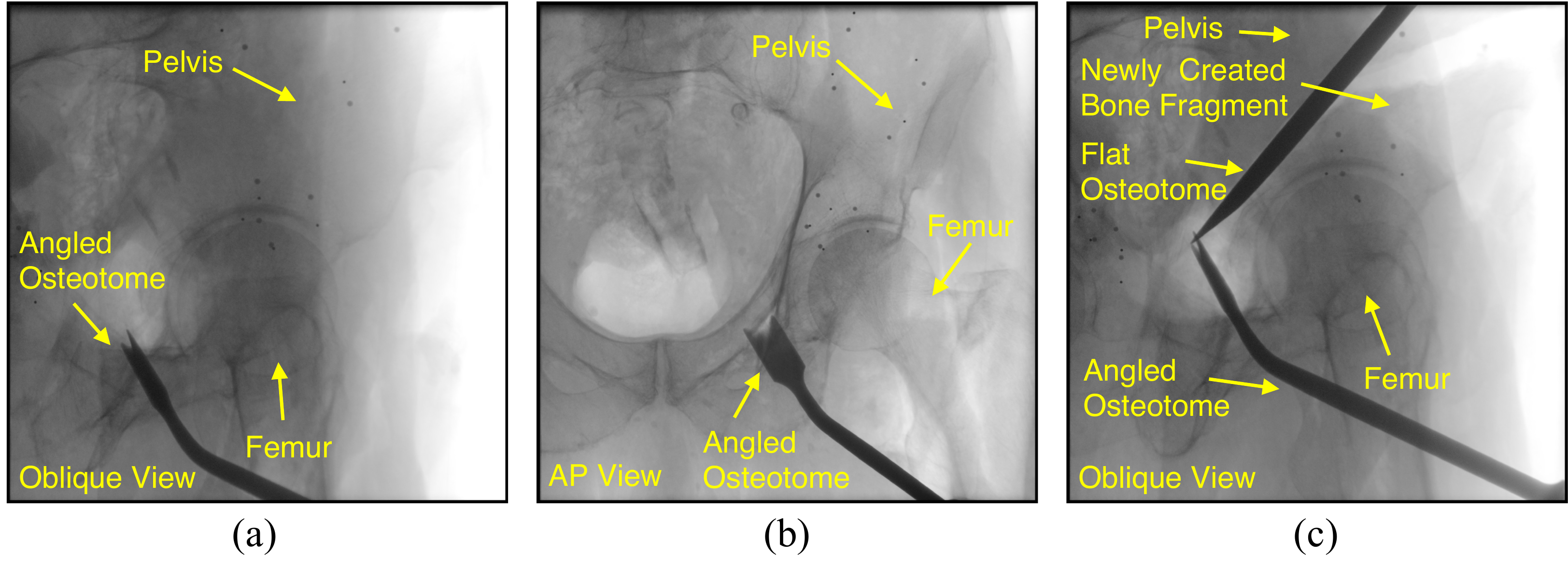}
    \caption{Fluoroscopic images of chiseling performed during a cadaveric periacetabular osteotomy procedure.
             A cut along the ischium of the pelvis is shown by the oblqiue view in (a).
             Due to the difficulty with manually interpreting lateral orientations of the tool in an oblique view, the very next fluoroscopy frame was collected at an approximate anterior-posterior (AP) view, shown in (b).
             Although the osteotome location was confirmed by changing view points, the process of adjusting the C-arm by such a large offset can potentially increase operative time or cause the clinician to lose some context in the previous oblique view.
             Another oblique view is shown in (c), where the flat osteotome is used to complete the posterior cut, resulting in the creation of two bone fragments from the pelvis.
             In (c), the angled osteotome was left in the field of view and used as a visual aid for navigating the flat chisel and completing the osteotomy.}
    \label{fig:pao_chisel}
\end{figure}

Image-guidance systems often need to process and report information regarding the relative poses between several objects of interest, such as bones, organs and surgical instruments. 
We found that only three studies in our review register multiple objects through learning-based approaches~\cite{varnavas2013fully,varnavas2015fullyB,wang2020multi}, and moreover, these studies independently registered single objects in order to obtain relative poses. 
Although combining the results of two distinct single object registrations is perhaps the most straight-forward approach for obtaining the relative pose between two objects, it fails to leverage the combined information of their relative pose during any optimization process, which could have potentially yielded a less-challenging search landscape. 
As an example, consider the case of two adjacent objects whose independent, single view, depth estimates are erroneous in opposite directions.
The relative pose computed from these two independent poses will have an exacerbated translation error resulting from the compounding effect of the independent depth errors.
A multiple object registration strategy could alternatively parameterize the problem to \emph{simultaneously} solve for the pose of one object with respect to the imaging device and also for the relative pose between the objects.
This approach partially compounds motion of the objects during registration and completely eliminates the possibility of conflicting depth estimates.\\
Despite the small number of learning-based multiple object registration strategies, traditional intensity-based registration methods are now routinely employed to solve compound, multiple object, registration problems across broad applications, such as:
kinematic measurements of bone~\cite{chen2012automatic,otake2016robust,abe2019analysis},
rib motion and respiratory analysis~\cite{hiasa2019recovery},
intra-operative assessment of adjusted bone fragments~\cite{grupp2019pose,grupp2020fast,han2021fracture},
confirmation of screw implant placement during spine surgery~\cite{uneri2017intraoperative} and
the positioning of a surgical robot with respect to target anatomy~\cite{gao2020fiducial_tmrb}.
We believe that the lack of new multiple object registration learning-based strategies is indicative of the substantial challenges involved with their development, rather than any perceived lack of the problem's importance by the community.
In order to better understand this ``gap'' between learning-based and traditional intensity-based methods in the multiple object domain, we first revisit \eqref{eq:reg_objective} and update it to account for $N$ 3D objects.
For $i = 1, \dots, N$, let $\omega_{D}^{(i)}$ denote the deformation parameters of the $i^{\text{th}}$ object and let $^{v}T_{x^{(i)}}$ denote it's pose  with respect to the $v^{\text{th}}$ view.
Since the vast majority of studies examined concern 2D X-ray images, we also assume that 2D view modality is X-ray.
This allows us to take advantage of the line integral nature of X-ray projection physics and represent the synthetic X-ray images formed from multiple objects as the sums of independent synthetic images created from each individual object, yielding the updated registration objective function:
\begin{equation}
    \{\hat{K}_v, \,
       ^{v}\hat{T}_{x^{(i)}},
       \hat{\omega}_{D}^{(i)}
    \}
       =
       \underset{K_v, \,^{v}T_{x^{(i)}},\omega_{D}^{(i)}}{\argmin}
            \sum_v S\!\left( \sum_{i} \left[ (P(K_v, \,^{v}T_{x^{(i)}}) \circ D_{\omega_{D}^{(i)}})(x^{(i)}) \right],
                             y_v  \right) + R(\cdot)\,.
    \label{eq:reg_objective_mult_obj}
\end{equation}
Solving this optimization becomes more challenging as objects are added and the dimensionality of the search space grows.
These challenges are somewhat mitigated by the compositional nature of the individual components of~\eqref{eq:reg_objective_mult_obj}.
Indeed, updating an intensity-based registration framework to compute \eqref{eq:reg_objective_mult_obj} instead of \eqref{eq:reg_objective} is relatively straight forward from an implementation perspective: compute $N$ synthetic radiographs instead of one and perform a pixel-wise sum of the synthetic images together before calculating the image similarity metric, $S(\cdot,\cdot)$.
The compositional structure of~\eqref{eq:reg_objective_mult_obj} also enables the high-dimensional problem to be solved by successively solving lower-dimensional sub-problems.
For example, the pose of a single object may be optimized while keeping the poses of all other objects fixed at their most recent estimates.
After this optimization of a single object is complete, another object's pose is optimized and all other's are kept constant.
This process is cycled until all objects have been registered once or some other termination criteria is met.\\
Extending~\eqref{eq:reg_objective_mult_obj} to the ML case, requires new per-object model parameters to be introduced: $\theta_D^{(i)}$ and $\theta_x^{(i)}$ for $i = 1, \dots, N$.
The updated ML-based objective function for multiple objects is written as:
\begin{equation}
    \{\hat{K}_v, \,
      ^{v}\hat{T}_{x^{(i)}},
      \hat{\omega}_{D}^{(i)}
    \}
        =
        \underset{K_v, \,^{v}T_{x^{(i)}},\omega_{D}^{(i)}}{\argmin}
            \sum_v S^{\mlemph{\theta_S}}\!\left(
                \sum_{i} \left[
                (P(K_v, \,^{v}T_{x^{(i)}}) \circ D^{\mlemph{\theta_D^{(i)}}}_{\omega_{D}^{(i)}})(\mlemph{G_x^{\theta_x^{(i)}}(}x^{(i)}\mlemph{)})
                \right],
                \mlemph{G_y^{\theta_y}(}y_v\mlemph{)}
                \right) 
                + R^{\mlemph{\theta_R}}(\cdot)\,.
    \label{eq:reg_objective_mult_obj_ML}
\end{equation}
When considering the $G_x^{\theta_x^{(i)}}(x^{(i)})$ terms in~\eqref{eq:reg_objective_mult_obj_ML}, it appears that contextualization methods are able to isolate some object-dependent parameters from other components of the registration problem.
However, we found that each of the contextualization studies which considered multiple objects~\cite{doerr2020data,esfandiari2021deep,grupp2020automatic,wang2020multi} did not explicitly isolate the parameters corresponding to each object, but instead used a single NN with data in the output layer indexed according to the appropriate object.
This is likely effective for very specific and simple applications as the NN is able to learn the relative spatial relationships of the objects.
For more complex applications, it may be more appropriate to learn separate, well-trained, models which may be composed together.
As an example, consider two separate models, one trained to identify contextual information of the hip joint and the other trained to produce contextual information of a robotic device.
Although these two models may be used in conjunction, or combined through ``fine-tuning,'' for the application of robotic hip surgery, the models may also be composed with models developed for other applications, such as measuring hip biomechanics or robotic shoulder surgery.
However, a singular model trained to jointly contextualize hip and robot features would most likely fail to generalize to any additional applications.
We anticipate that the development of independent and robust contextualization models, capable of composition, shall accelerate the number learning-based methods applied to the multiple object problem.\\
Although representation learning methods may function in the presence of multiple objects without additional modification, better performance is usually obtained by adding structure to account for the known spatial relationships between objects~\cite{yokota2013automated}.
We anticipate that methods relying on representation learning shall migrate away from PCA, towards auto-encoder style approaches, with embedded rigid transformer modules so that all spatial and shape relationships may be learned.
As representation learning methods have the potential to reconstruct the bony anatomy of joints from a sparse number of 2D views, these could potentially become very popular by enabling navigation to a 3D anatomical model with substantially reduced radiation exposure.\\
Similarity modeling approaches are implicitly affected by the introduction of additional objects due to the inputs of $S^{\theta_S}(\cdot,\cdot)$ having a dependence on $G_x^{\theta_x^{(i)}}(x^{(i)})$ and $\theta_D^{(i)}$.
None of the studies reviewed in this paper used multiple objects as part of similarity modeling.
Although~\cite{grupp2020automatic} registered multiple objects, the regularization function employed learned landmark annotations from a only a single object.
We envision several challenges with extending these methods to properly accommodate multiple objects.
Obtaining convexity of the learned similarity models in the ideal case for~\cite{gao2020generalizing,gu2020extended} is most likely not attainable when using a weighted sum of rotation and translation components of each object's pose offset, especially as the number of objects increases.
Methods which model 2D/3D correspondences~\cite{schaffert2019metric,schaffert2020learning,schaffert2020learningB,liao2019multiview} will require an additional dimension to handle the assignment of points to various obejcts, adding complexity and potentially increasing the challenges associated with training.\\
Probably most handicapped by the introduction of multiple objects to the registration problem, are methods relying on the direct regression of pose parameters, as they attempt to model the entire objective function.
One may be tempted to solve this problem by simply adding additional model outputs for each object's estimated pose, but this does not guarantee that the limitations of independent single object pose regression are addressed.
Therefore, direct multiple object pose regression methods should attempt to model the relative poses between objects in addition to a single absolute pose (or single pose relative to some initialization).
Another limitation of regression approaches lies in the combinatorial explosion which occurs as new objects are added to the registration problem, making training difficult in the presence of large fluctuations of the loss function.\\
Although not unique to learning-based methods, multiple object registration verification also becomes a much more complex problem as the number of objects considered increases.
Questions arise, such as: should verification be reported on each relative pose or should an overall pass/fail be reported?
The three verification studies examined in this paper only considered verification of a single object's pose estimate.
As single object verification methods mature, their issues when expanding to multiple objects will likely become more apparent.\\
In light of these challenges associated with learning-based approaches, it is easy to see how contemporary intensity-based methods currently dominate the multiple object domain given their relative ease of implementation and reasonable performance.
However, there are multiple object registration problems which remain unsolved since
multiple object intensity-based methods continue to suffer from the limitations previously identified in Section~\ref{subsec:problem}.
Some frequent properties of these unsolved problems are misleading views with several objects having substantial overlap in 2D,
the potential for a varying number of surgical instruments to be present in a view
and the existence of objects which are dynamically changing shape or possibly ``splitting'' into several new objects.
Real-time osteotome navigation using single view fluoroscopy, illustrated in Figure~\ref{fig:pao_chisel}, is exemplary of the many unsolved challenges outlined above and will be used as a motivating example throughout this discussion.\\
Each time the osteotome is advanced through bone, a single fluoroscopic view is collected and interpreted by the surgeon in order to accurately adjust the chiseling trajectory and safely avoid sensitive components of the anatomy which must not be damaged, such as the acetabulum shown in~\ref{fig:pao_chisel}.
Although the oblique views shown in Figure~\ref{fig:pao_chisel} (a) and (c) help ensure that the osteotome tip is distinguishable from the acetabulum, the lateral orientation of the tool is challenging to interpret manually and may need to be confirmed by collecting subsequent fluoroscopic views at substantially different orientations, such as the approximate anterior-posterior (AP) view shown in Figure~\ref{fig:pao_chisel} (b).
Figure~\ref{fig:pao_chisel} (c) also demonstrates the intraoperative creation of a new object which may move independently of all others, further compounding the difficulty associated with non-navigated interpretation of these views.
An additional challenge with osteotomy cases is the ability to determine when all osteotomies are complete and the new bone fragment is completely free from its parent.
Figure~\ref{fig:pao_chisel} (c) demonstrates that retaining the angled osteotome as a static object in the field of view during performance of the posterior cut helps to guide the flat osteotome along the desired trajectory, but does not guarantee that the acetabular fragment has in fact been freed from the pelvis.
A navigation system capable of accurately tracking the osteotome with respect to the anatomy using a series of oblique views would eliminate the need for additional, ``confirmation,'' views, potentially reduce operative time, decrease radiation exposure to the patient and clinical team and reduce the frequency of breaches into sensitive anatomy.\\
We envision that this problem, and others like it, will eventually be solvable as the capabilities of learning-based methods advance further.
Future contextualization and similarity modeling methods could enable a large enough capture range to provide an automatic registration to the initial oblique view used for chiseling, and regression methods, coupled with contextualization, could enable real-time pose estimates in subsequent views.
Reconstructions of newly created bone fragments should be possible through similarity modeling approaches, such as by extending the framework introduced by~\cite{gao2020generalizing} to include multiple objects and non-rigid deformation.
Finally, recurrent NN approaches, such as long short-term memory (LSTM) components, could provide some \emph{temporal} segmentation of the intervention into phases and gestures, as is already done for laparoscopic surgery~\cite{vercauteren2019cai4cai,garrow2021machine,wu2021cross}.
This segmentation could be useful in identifying \emph{when} 1) certain objects need to be tracked, 2) a radically different view has been collected, or 3) new objects have been split off from an existing object.\\
Although foreign objects such as screws, K-wires, retractors, osteotomes, etc. frequently confound traditional registration methods, their location or poses often have clinical relevance.
Therefore, contrary to the inpainting approach of~\cite{esfandiari2021deep}, we advocate that learning-based methods should make attempts to register these objects.

\subsection{Estimating uncertainty and assuring quality}
The four studies which consider registration verification in this review all attempt to provide a low dimensional classification of registration success, e.\,g. pass/fail or correct/poor/incorrect~\cite{mitrovi2014automatic,varnavas2013fully,varnavas2015fully,wu2016a}.
These strategies effectively attempt to label registration estimates as either a global (correct) or local minimum (poor/incorrect) of~\eqref{eq:reg_objective}.
Unfortunately, even when correctly classifying a global optimum, these low-dimensional categorizations of a registration result may unintentionally fail to report small, but clinically relevant, errors. 
This is perhaps easiest to recognize by first revisiting~\eqref{eq:reg_objective} and noting that registration strategies attempt to find singular solutions, or \emph{point} estimates, which best minimize the appropriate objective function.
However, several factors, including sensor noise, modeling error or numerical imprecision, may influence the landscape of the objective function and potentially even cause some variation in the location of the global minimum.
This possibility of obtaining several different pose estimates under nominally equivalent conditions, reveals the inherent \emph{uncertainty} of registration.
Even though a failure to identify cases of large uncertainties may lead surgeons to take unintentional risks with potentially catastrophic implications for patients, to our knowledge, only one prior work has attempted to estimate the error associated with 2D/3D registration estimates~\cite{hu20162d}.
We therefore believe that the development of methods which quantify 2D/3D registration uncertainty is of paramount importance and essential for the eventual adoption of 2D/3D registration into routine use, and even more in the advent of autonomous robotic surgery.\\
Inspiration can be drawn from the applications of 3D/3D and 2D/2D deformable image registration~\cite{chen2020deep,haskins2020deep,fu2020deep}, where machine learning techniques have dominated most existing research into registration uncertainty.
Some of these approaches report \emph{interval} estimates, either by reformulating the registration objective function as a probability distribution and drawing samples~\cite{risholm2013bayesian,le2016quantifying,schultz2018multilevel}, approximating the sampling process using test-time NN drop-out~\cite{yang2017quicksilver} or sampling using the test-time deformation covariance matrices embedded within a variational autoencoder~\cite{dalca2019unsupervised}.
Due to the interventional nature of most 2D/3D registration applications, care needs to be taken in order to ensure that program runtimes are compatible with intra-operative workflows.\\
Although we have mostly advocated for the development of new uncertainty quantification methods, there is likely still room for improvement of the local/global minima classification problem.
Given the computational capacity of contemporary GPUs and sophistication of learning frameworks, it should be feasible to extend the approach of~\cite{wu2016a} and pass densely sampled regions of the objective function through a NN, relying on the learning process to extract optimal features for distinguishing local and global minimal.\\
As registration methods will inevitably report failure or large uncertainties under certain conditions, we also believe that a promising topic of future research extends to intelligent agents which would determine subsequent actions to optimally reduce uncertainty, such as collecting another 2D view from a specific viewpoint.
Finally, we note that registration uncertainties have the potential to augment existing robotic control methods which rely on intermittent imaging~\cite{alambeigi2019scade}.

\subsection{The need for more generic solutions}
Traditional image-based 2D/3D registration that relies on optimization of a cost function is limited in many ways (Section~\ref{subsec:problem}); however, a major advantage of the algorithmic approach is that it is very generic, i.\,e., a pipeline configured for 2D/3D registration of the pelvis would be equally applicable to the spine. The introduction of machine learning to the 2D/3D pipeline, however, has in a way taken away this generality and methods have become substantially more specialized. This is not because specific machine learning models are only suitable for one specific task or anatomy; on the contrary, e.\,g., pose or PCA component regression techniques are largely identical across clinical targets. Rather, it is yet another consequence of the compact domain assumption of machine learning models that confine solutions to the data domain they were developed on. While this specificity of solutions may be acceptable, and perhaps, unavoidable for methods that seek to contextualize data, it inhibits the use of 2D/3D registration pipelines as a tool to answer clinical or research questions. We feel that, in part due to its complexity, 2D/3D registration already leads a niche existence and the necessity to re-train or even re-develop some algorithm components for the specific task at hand further exacerbates this situation.\\
Some approaches already point towards potential solutions to this issue: Methods like \cite{varnavas2013fully,varnavas2015fullyB} rely on patient- and object-specific training sets, and therefore, avoid the task-specificity that comes with one-time training. Another noteworthy method learns to establish sparse point correspondences between random points rather than anatomically relevant ones~\cite{liao2019multiview}. However, despite matching random points, this method is likely still scene-specific because matching is performed using a very deep CNN with very large receptive field which may have learned to exploit the global context of the random keypoints, which would not be preserved in different anatomy. Contributing machine learning-based solutions that address some of the large open challenges in 2D/3D registration while making the resulting tools general purpose and easy to use will be an important goal in the immediate future.

\section{Conclusion}
Machine learning-based improvements to image-based 2D/3D registration were already of interest before the deep learning era~\cite{gouveia2012comparative}, and deep learning has only accelerated and diversified the contributions to the field. Contextualization of data, representation learning to reduce problem dimensionality, similarity modeling for increased capture range, direct pose regression to avoid iterative optimization, as well as confidence assessment are all well established research thrusts, which are geared towards developing automated registration pipelines. While convincing performance improvements are reported across a variety of clinical tasks and problem settings already today, most of those studies are performed ``on the benchtop.''\\
Coordinated research efforts are desirable towards 1) developing more robust learning paradigms that succeed under domain shift, 2) creating standardized reporting templates and devising evaluation metrics to enhance reproducibiliy and enable comparisons, 3) researching multi-object registration methods that can deal with the presence of foreign objects, 4) advancing uncertainty quantification and confidence estimation methodology to better support human decisions, and finally 5) developing generalist machine learning components of 2D/3D registration pipelines to improve accessibility.\\
Even though learning-based methods show great promise and have supplanted traditional methods in many aspects, their rise should not render traditional methods unusable or irrelevant. New researchers typically spend a great deal of time implementing a traditional registration pipeline so that traditional methods may be leveraged in conjunction with the development of learning-based approaches. 
In order to facilitate more rapid research and development towards learning-based methods, we have made the core registration components of our intensity-based framework (xReg\footnote{\href{https://github.com/rg2/xreg}{https://github.com/rg2/xreg}}), as well as our physics-based DRR generation tools for realistic synthesis and generalizable learning (DeepDRR\footnote{\href{https://github.com/arcadelab/deepdrr}{https://github.com/arcadelab/deepdrr}}) available as open source software projects.\\
We have no doubt that progress on the aforementioned fronts will firmly establish machine learning methodology as an important component for 2D/3D registration workflows that will substantially contribute to 2D/3D registration growing out of its niche existence, establishing itself as a reliable, precise, and easy-to-use component for research, and more importantly, at the bedside.


\section*{Conflict of Interest Statement}
The authors declare that the research was conducted in the absence of any commercial or financial relationships that could be construed as a potential conflict of interest.

\section*{Author Contributions}
M.U. conceived of the presented idea was in charge of overall direction and planning. 
M.U., C.G., and R.G. refined the scope of the presented review and perspective. 
M.U., C.G., R.G., M.J, and Y.H. screened abstracts and full texts during review and extracted information from the included studies. 
M.U., C.G., and R.G. merged extracted information and carried out the systematic review. 
All authors contributed to writing the manuscript. They also provided critical feedback, and helped shape the research and analysis.

\section*{Funding}
We gratefully acknowledge financial support from NIH NIBIB Trailblazer R21 EB028505, and internal funds of the Malone Center for Engineering in Healthcare at Johns Hopkins University.

\section*{Acknowledgments}
The viewpoints presented in this manuscript have been shaped by several years of work on 2D/3D registration in the context of image-guided surgery at Johns Hopkins University. Therefore, the many people who contributed to that body of work also had an influence on how we think about research on this topic today -- Thank you.

\section*{Data Availability Statement}
There is no data associated with this article.




\tiny
\begin{longtable}{|p{1.5cm}|p{2.7cm}|p{2.5cm}|p{2.5cm}|p{2.5cm}|p{2.5cm}|}
\caption{\revised{A summary of each study's relation with the five themes of Contextualization, Representation learning, Direct parameter regression, Similarity modeling, and Verification.  }}
\label{table:5themes}
\\
\hline
\textbf{Study ID} & \textbf{Contextualization} & \textbf{Representation learning} & \textbf{Direct parameter regression} &\textbf{Similarity modeling}  & \textbf{Verification}
\\
\hline
\endfirsthead
\caption*{Table~\ref{table:5themes} Continued}
\\
\hline
\textbf{Study ID} & \textbf{Contextualization} & \textbf{Representation learning} & \textbf{Direct parameter regression} &\textbf{Similarity modeling}   & \textbf{Verification}
\\
\hline
\endhead
\cite{brost2012constrained} &\cellcolor{Apricot} &\cellcolor{blue!25} & & & \\\hline
\cite{lin2012shape} &\cellcolor{Apricot} & & & & \\\hline
\cite{chou2013real} & &\cellcolor{blue!25}&\cellcolor{SkyBlue} & & \\\hline
\cite{chou20132d} & &\cellcolor{blue!25}&\cellcolor{SkyBlue}  & & \\\hline
\cite{varnavas2013fully} &\cellcolor{Apricot} & & & &\cellcolor{YellowGreen} \\\hline
\cite{chou2014local} & &\cellcolor{blue!25}&\cellcolor{SkyBlue} & & \\\hline
\cite{mitrovi2014automatic} & & & & &\cellcolor{YellowGreen} \\\hline
\cite{zhao2014local} & &\cellcolor{blue!25}&\cellcolor{SkyBlue} & & \\\hline
\cite{baka2015respiratory} &\cellcolor{Apricot} &\cellcolor{blue!25} & & & \\\hline
\cite{mitrovi2015simultaneous} & & &\cellcolor{SkyBlue} & & \\\hline
\cite{varnavas2015fully} &\cellcolor{Apricot} & & & &\cellcolor{YellowGreen} \\\hline
\cite{wu2015fully} & & &\cellcolor{SkyBlue} & & \\\hline
\cite{miao2016real} & & &\cellcolor{SkyBlue} & & \\\hline
\cite{miao2016a} & & &\cellcolor{SkyBlue} & & \\\hline
\cite{tang2016similarity} & & & &\cellcolor{purple!90} & \\\hline
\cite{wu2016a} & & & & &\cellcolor{YellowGreen} \\\hline
\cite{hou2017predicting} & & &\cellcolor{SkyBlue} & & \\\hline
\cite{pei2017non} & &\cellcolor{blue!25}&\cellcolor{SkyBlue} & & \\\hline
\cite{xie2017single} & & &\cellcolor{SkyBlue} & & \\\hline
\cite{bier2018x} &\cellcolor{Apricot} & & & & \\\hline
\cite{chen2018real} &\cellcolor{Apricot} &\cellcolor{blue!25} & & & \\\hline
\cite{hou2018computing} & & &\cellcolor{SkyBlue} & & \\\hline
\cite{miao2018dilated} & & &\cellcolor{SkyBlue} & & \\\hline
\cite{toth20183d} & & &\cellcolor{SkyBlue} & & \\\hline
\cite{zhang2018temporal} & &\cellcolor{blue!25} &\cellcolor{SkyBlue} & & \\\hline
\cite{zheng2018pairwise} & & &\cellcolor{SkyBlue} & & \\\hline
\cite{bier2019learning} &\cellcolor{Apricot} & & & & \\\hline
\cite{foote2019real} & &\cellcolor{blue!25}&\cellcolor{SkyBlue} & & \\\hline
\cite{guan2019transfer} & & &\cellcolor{SkyBlue} & & \\\hline
\cite{liao2019multiview} & & & &\cellcolor{purple!90} & \\\hline
\cite{luo2019towards} &\cellcolor{Apricot} & & & & \\\hline
\cite{schaffert2019metric} & & & &\cellcolor{purple!90} & \\\hline
\cite{yang2019a} &\cellcolor{Apricot} & & & & \\\hline
\cite{doerr2020data} &\cellcolor{Apricot} & & & & \\\hline
\cite{francoise2020detecting} &\cellcolor{Apricot} & & &\cellcolor{purple!90} & \\\hline
\cite{gao2020generalizing} & & &\cellcolor{SkyBlue} &\cellcolor{purple!90} & \\\hline
\cite{grupp2020automatic} &\cellcolor{Apricot} & & &\cellcolor{purple!90} & \\\hline
\cite{gu2020extended} & & & &\cellcolor{purple!90} & \\\hline
\cite{guan2020transfer} & & &\cellcolor{SkyBlue} & & \\\hline
\cite{karner2020single} &\cellcolor{Apricot} & & & & \\\hline
\cite{li2020non} & &\cellcolor{blue!25}&\cellcolor{SkyBlue} & & \\\hline
\cite{neumann2020deep} & & & &\cellcolor{purple!90} & \\\hline
\cite{schaffert2020learning} & & & &\cellcolor{purple!90} & \\\hline
\cite{schaffert2020learningB} & & & &\cellcolor{purple!90} & \\\hline
\cite{wang2020multi} &\cellcolor{Apricot} & & & & \\\hline
\cite{xiangqian20202d} & & &\cellcolor{SkyBlue} & & \\\hline
\cite{zhang2020automatic} & &\cellcolor{blue!25} & & & \\\hline
\cite{esfandiari2021deep} &\cellcolor{Apricot} & & & & \\\hline
\end{longtable}

\tiny
\begin{longtable}{|p{1.5cm}|p{1.5cm}|p{1.7cm}|p{2.2cm}|p{2.2cm}|p{2.2cm}|p{0.9cm}|p{1.3cm}|}
\caption{Parameters defining the registration problems described in the studies included for review. Registration purpose refers to the registration stage being addressed, such as initialization (init.), precise retrieval of geometric parameters (fine regis.), or others.}
\label{table:registration parameter}
\\
\hline
\textbf{Study ID} & \textbf{3D Modality} & \textbf{2D Modality} & \textbf{Regis. Action} & \textbf{Regis. Purpose} & \textbf{Speciality} & \textbf{Rigid / Non-rigid} & \textbf{No. Views}
\\
\hline
\endfirsthead
\caption*{Table~\ref{table:registration parameter} Continued}
\\
\hline
\textbf{Study ID} & \textbf{3D Modality} & \textbf{2D Modality} & \textbf{Regis. Action} & \textbf{Regis. Purpose} & \textbf{Speciality} & \textbf{Rigid / Non-rigid} & \textbf{No. Views}
\\
\hline
\endhead
\cite{brost2012constrained} &Catheter Model&X-ray &Pre-proc&fine regis.&Catheter&R&1 \\\hline
\cite{lin2012shape} &CT&X-ray&Pre-proc&fine regis.&Radiotherapy&R&2 \\\hline
\cite{chou2013real}&CT (3D+t)&X-ray&Deformable regr.&fine regis.&Lung tumors&NR&1 \\\hline
\cite{chou20132d}&CBCT, s-NST&X-ray&Pose updates&fine regis.&Head-and-neck, lungs&Both&R: 4, NR: 2 \\\hline
\cite{varnavas2013fully}&CT&X-ray&Pre-proc&init., verif.&Spine/Vertebrae&R&1 \\\hline
\cite{chou2014local}&CT&X-ray&Deformable regr.&fine regis.&Lung&NR&1 \\\hline
\cite{mitrovi2014automatic}&CT&X-ray&N/A&verif.&Spine/Vertebra&R&1 \\\hline
\cite{zhao2014local}&CT&X-ray&Deformable regr.&fine regis.&Abdomen&NR&1 \\\hline
\cite{baka2015respiratory}&CTA&XA&Pre-proc&fine regis.&Coronary artery&Both&1 (temporal) \\\hline
\cite{mitrovi2015simultaneous}&DSA&DSA&Pose regr., updates&init., fine regis.&Angiography&R&1 \\\hline
\cite{varnavas2015fully}&CT&X-ray&Pre-proc&init., verif.&Spine&R&1 \\\hline
\cite{wu2015fully}&mdl.&X-ray&Pose regr.&init.&Knee&R&1 \\\hline
\cite{miao2016real}&Tool Model&X-ray&Pose regr.&fine regis.&Implants&R&1 \\\hline
\cite{miao2016a}&Implant and TEE mdl.&X-ray&Pose regr.&fine regis.&Implants / TEE transducer&R&1 \\\hline
\cite{tang2016similarity}&MRA&DSA&cost func&fine regis.&Angiography&R&1 \\\hline
\cite{wu2016a}&CT&X-ray&N/A&verif.&Skull/Head&R&2 \\\hline
\cite{hou2017predicting}&CT&X-ray&Pose regr. &init.&Thorax&R&1 \\\hline
\cite{pei2017non}&CT&X-ray&Deformable regr.&fine regis.&Skull&NR&1 \\\hline
\cite{xie2017single}&CTA&X-ray&Pose regr.&fine regis.&Angiography&R&1 \\\hline
\cite{bier2018x}&CT&X-ray&Pre-proc&init.&Pelvis&R&1 \\\hline
\cite{chen2018real}&CT&X-ray&Pre-proc&feat. extract.&Spine&R&1 \\\hline
\cite{hou2018computing}&CT&X-ray&Pose regr.&init.&Thorax &R&1 \\\hline
\cite{miao2018dilated}&CBCT&X-ray&Pose regr., updates&fine regis.&Spine&R&2 \\\hline
\cite{toth20183d}&Left ventr mdl.&X-ray&Pose regr., updates&fine regis.&Heart&R&1 \\\hline
\cite{zhang2018temporal}&PCA Deformation Field&X-ray&Deformable regr.&init., fine regis.&Skull&NR&1 ( temporal ) \\\hline
\cite{zheng2018pairwise}&CT, implant mdl.&X-ray&Domain adaptation&Pose regr., updates&Spine and TEE transducer&R&1 \\\hline
\cite{bier2019learning}&CT&X--ray&Pre-proc &init.&Pelvis&R&1 \\\hline
\cite{foote2019real}&CT&X-ray&Deformable  regr.&fine regis.&Lung&NR&1 \\\hline
\cite{guan2019transfer}&Vasc mdl. (CT)&DSA&Pose regr.&fine regis.&Cardiovascular&R&1 \\\hline
\cite{liao2019multiview}&CT or CBCT&X-ray&cost func&fine regis.&Thorax&R&1 \\\hline
\cite{luo2019towards}&CT&Broncho - scopy&Pre-proc&fine regis.&Bronchoscopy&R&1 \\\hline
\cite{schaffert2019metric}&CBCT&X-ray&cost func&fine regis.&Spine&R&1 \\\hline
\cite{yang2019a}&CT/MRI&Stereo RGB&Pre-proc&feat. extract.&Head/face&R&2 \\\hline
\cite{doerr2020data}&N/A &X-ray&Pre-proc&init.&Spine, pedicle screws&R&1 \\\hline
\cite{francoise2020detecting}&MRI&Laparo - scopy&Pre-proc, cost func&fine regis.&Uterus&N/A&N/A \\\hline
\cite{gao2020generalizing}&CT&X-ray&Pose regr.&Pose updates &Pelvis&R&1 \\\hline
\cite{grupp2020automatic}&CT&X-ray&Pose regr.&Pose updates &Pelvis&R&1 \\\hline
\cite{gu2020extended}&CT&X-ray&cost func&Pose updates&Pelvis&R&1 \\\hline
\cite{guan2020transfer}&Aorta mdl.&DSA (synth)&Deformable regr. &fine regis.&Cardiovascular&NR&1 \\\hline
\cite{karner2020single}&CT/MR&RGB face img&Pre-proc&fine regis.&Face&R&1 \\\hline
\cite{li2020non}&CBCT&X-ray&Deformable regr. &fine regis.&Skull&NR&1 \\\hline
\cite{neumann2020deep}&MRA&DSA &cost func&fine regis.&Angiography&R&1 \\\hline
\cite{schaffert2020learning}&CBCT&X-ray&cost func&fine regis.&Spine&R&1 \\\hline
\cite{schaffert2020learningB}&CBCT&X-ray&cost func&fine regis.&Spine, head&R&1-2 \\\hline
\cite{wang2020multi}&CT&Bi - plane Fluoroscopy&Pre-proc&init.&Knee&R&2 \\\hline
\cite{xiangqian20202d}&CT&X-ray&Pose regr.&fine regis.&Pelvis&R&1 \\\hline
\cite{zhang2020automatic}&CT&X-ray&Post-proc&fine regis.&Liver tumors&NR&20 \\\hline
\cite{esfandiari2021deep} &CT&X-ray&Pre-proc &fine regis.&Spine, pedicle screws&R&2\\\hline
\end{longtable}

\begin{longtable}{|p{1.5cm}|p{1.8cm}|p{1.3cm}|p{2cm}|p{1.5cm}|p{1.7cm}|p{2cm}|p{1.7cm}|}
\caption{A summary of the training and testing details for each study reviewed.}
\label{table:training/testing}
\\
    \hline
    \multicolumn{1}{|c}{} &
    \multicolumn{1}{|c}{\textbf{Network}} &
    \multicolumn{5}{|c}{\textbf{Training}} &
    \multicolumn{1}{|c|}{\textbf{Testing}} \\

    \hline
    \textbf{Study ID} & \textbf{Architecture} & \textbf{Data} & \textbf{Domain Transfer} & \textbf{Object Number} & \textbf{Anatomy Specificity} & \textbf{Technique} & \textbf{Data}
    \\
    \hline
    \endfirsthead
    \caption*{Table~\ref{table:training/testing} Continued}
    \\
    \hline
    \multicolumn{1}{|c}{} &
    \multicolumn{1}{|c}{\textbf{Network}} &
    \multicolumn{5}{|c}{\textbf{Training}} &
    \multicolumn{1}{|c|}{\textbf{Testing}} \\

    \hline
    \textbf{Study ID} & \textbf{Architecture} & \textbf{Data} & \textbf{Domain Transfer} & \textbf{Object Number} & \textbf{Anatomy Specificity} & \textbf{Technique} & \textbf{Data}
    \\
    \hline
    \endhead
    \cite{brost2012constrained}&PCA&Real&Segmentation&Single&N/A&Unsupervised&Real \\
    \hline
    \cite{lin2012shape}&N/A&N/A&N/A&Single&Patient&Unsupervised&N/A \\
    \hline
    \cite{chou20132d}&Linear Regression&Synthetic&N/A&Single&Anatomy&Supervised&Synthetic \\
    \hline
    \cite{chou2013real}&PCA&Synthetic and Real&N/A&Single&Anatomy&Supervised&Synthetic and Real \\
    \hline
    \cite{varnavas2013fully}&GHT&Real&N/A&Single&Anatomy and Patient&Unsupervised&Real\\
    \hline
    \cite{chou2014local}&Random Forest&Synthetic&Gaussian Normalization&Single&Patient&Supervised&Synthetic and Real \\
    \hline
    \cite{mitrovi2014automatic}&N/A&Synthetic&N/A&Single&Patient&Supervised&Real \\
    \hline
    \cite{zhao2014local}&Linear Regression&Synthetic&N/A&Single&Patient&Supervised&Real and Synthetic \\
    \hline
    \cite{baka2015respiratory}&N/A&Real&N/A&Single&Patient&N/A&Real \\
    \hline
    \cite{mitrovi2015simultaneous}&N/A&N/A&N/A&Single&Patient&N/A&Synthetic and Real \\
    \hline
    \cite{varnavas2015fully}&GHT&Synthetic&N/A&Single&Anatomy and Patient&Unsupervised&Real\\
    \hline
    \cite{wu2015fully}&PCA&Synthetic&N/A&Single&Anatomy&Unsupervised&Synthetic and Real\\
    \hline
    \cite{miao2016a}&Siamese CNNs&Synthetic&Realism Tuning&Single&Anatomy&Supervised&Real\\
    \hline
    \cite{miao2016a}&Siamese CNNs&Synthetic&N/A&Single&Anatomy&Supervised&Real\\
    \hline
    \cite{tang2016similarity}&Spectral Regression&Real&Abstraction&Single&Anatomy&Supervised&Real\\
    \hline
    \cite{wu2016a}&MLP&Real&N/A&Single&N/A&Supervised&Real\\
    \hline
    \cite{hou2017predicting}&CaffeNet&Synthetic&N/A&Single&Anatomy&Supervised&Synthetic\\
    \hline
    \cite{pei2017non}&CNNs&Real&N/A&Single&Anatomy&Supervised&Synthetic\\
    \hline
    \cite{xie2017single}&CNNs&Synthetic&N/A&Single&Anatomy&Supervised&Synthetic\\
    \hline
    \cite{bier2018x}&Sequential CNNs&Synthetic&Domain Generalization&Single&Anatomy&Supervised&Synthetic\\
    \hline
    \cite{chen2018real}&PCA&Synthetic&N/A&Single&N/A&Supervised&Real\\
    \hline
    \cite{hou2018computing}&N/A&Synthetic&N/A&Single&Anatomy&Supervised&Synthetic\\
    \hline
    \cite{miao2018dilated}&Dilated CNNs&Synthetic&N/A&Single&Anatomy&Supervised&Real\\
    \hline
    \cite{toth20183d}&CNNs&Synthetic&N/A&Single&Anatomy&Supervised&Synthetic\\
    \hline
    \cite{zhang2018temporal}&VGG&Synthetic&N/A&Single&Anatomy&Supervised&Synthetic\\
    \hline
    \cite{zheng2018pairwise}&DA Module&Pairwise Synthetic and Real&Domain Adaptation&Single&Anatomy&Unsupervised&Real\\
    \hline
    \cite{bier2019learning}&Sequential CNNs&Synthetic&Domain Generalization&Single&Anatomy&Supervised&Synthetic and Cadaver\\
    \hline
    \cite{foote2019real}&DenseNet&Synthetic&Equalization methods&Single&Patient&Supervised&Synthetic\\
    \hline
    \cite{guan2019transfer}&CNNs&Synthetic&Retrain on patient data&Single&Anatomy&Supervised&Synthetic\\
    \hline
    \cite{liao2019multiview}&U-Net&Real&N/A&Single&Anatomy&Supervised&Real\\
    \hline
    \cite{luo2019towards}&Instance Learning&Real&N/A&Single&Anatomy&Supervised&Real\\
    \hline
    \cite{schaffert2019metric}&PointNet&Real&N/A&Single&Anatomy&Supervised&Real\\
    \hline
    \cite{yang2019a}&Stacked Hourglass&Real&N/A&Single&Anatomy&Supervised&Synthetic\\
    \hline
    \cite{doerr2020data}&Fast R-CNN&Synthetic&N/A&Multiple&Patient&Supervised&Synthetic\\
    \hline
    \cite{francoise2020detecting}&U-Net&Real&N/A&Single&Anatomy&Supervised&Real\\
    \hline
    \cite{gao2020generalizing}&Spatial Transformer&Synthetic&Realistic Simulation&Single&Anatomy&Supervised&Synthetic and Real\\
    \hline
    \cite{grupp2020automatic}&U-Net&Real&N/A&Single&Anatomy&Supervised&Real\\
    \hline
    \cite{gu2020extended}&DenseNet&Synthetic&Realistic Simulation&Single&Anatomy&Supervised&Synthetic and Real\\
    \hline
    \cite{guan2020transfer}&CNNs&Synthetic&Refine on patient data&Single&Anatomy&Supervised&Synthetic\\
    \hline
    \cite{karner2020single}&Face-to-3D&N/A&N/A&Single&Anatomy&N/A&Real\\
    \hline
    \cite{li2020non}&ResNet&Synthetic&N/A&Single&Anatomy&Self-supervised&Synthetic and Real\\
    \hline
    \cite{neumann2020deep}&Siamese ResNet&Synthetic&N/A&Single&Anatomy&Supervised&Synthetic\\
    \hline
    \cite{schaffert2020learning}&PointNet&Real&N/A&Single&Anatomy&Supervised&Real\\
    \hline
    \cite{schaffert2020learningB}&FlowNet-S&Real&N/A&Single&Anatomy&Supervised&Real\\
    \hline
    \cite{wang2020multi}&VGGs&Real&Domain Adaptation&Multiple&Anatomy&Supervised&Real\\
    \hline
    \cite{xiangqian20202d}&GoogleNet&Synthetic&Histogram Matching&Single&Anatomy&Supervised&Real\\
    \hline
    \cite{zhang2020automatic}&U-Net&Real&N/A&Single&Anatomy&Supervised&Real\\
    \hline
    \cite{esfandiari2021deep}&PConvS&Synthetic&Heavy Augmentation&Single&Anatomy&Supervised&Synthetic\\
    \hline
\end{longtable}

\end{document}